\documentclass[journal,twoside,web]{IEEEtran}
\usepackage{generic}
\usepackage{cite}
\let\labelindent\relax
\usepackage{amsmath,amssymb,amsfonts, enumitem}
\usepackage{mathbbol}
\usepackage{bbm, dsfont}
\usepackage{graphicx}
\usepackage{textcomp}
\usepackage{mathtools}
\usepackage{algorithm, algpseudocode}
\usepackage{subcaption}
\usepackage{wrapfig}

\usepackage{booktabs}
\usepackage{tabularx}
\usepackage[T1]{fontenc}
\usepackage[utf8]{inputenc}
\newcolumntype{Y}{>{\centering\arraybackslash}X}

\newtheorem{theorem}{Theorem}[section]

\newtheorem{lemma}[theorem]{Lemma}
\newtheorem{corollary}[theorem]{Corollary}
\newtheorem{definition}[theorem]{Definition}
\newtheorem{assu}[theorem]{Assumption}
\newtheorem{remark}[theorem]{Remark}
\usepackage{algorithm}
\usepackage{algorithmicx}
\usepackage{algpseudocode}
\usepackage{setspace}
\usepackage{dsfont}
\usepackage{comment}

\usepackage{calrsfs}
\DeclareMathAlphabet{\pazocal}{OMS}{zplm}{m}{n}
\SetMathAlphabet\pazocal{bold}{OMS}{zplm}{bx}{n}

\makeatletter
\def\widebreve#1{\mathop{\vbox{\m@th\ialign{##\crcr\noalign{\kern\p@}%
  \brevefill\crcr\noalign{\kern0.1\p@\nointerlineskip}%
  $\hfil\displaystyle{#1}\hfil$\crcr}}}\limits}

\def\brevefill{$\m@th \setbox\z@\hbox{}%
 \hfill\scalebox{0.7}{\rotatebox[origin=c]{90}{(}} \kern4pt $}
\makeatletter

\newcommand{\nn}{\nonumber}
\newcommand{\X}{\pazocal{X}}
\newcommand{\U}{\pazocal{U}}
\newcommand{\V}{\pazocal{V}}
\renewcommand{\H}{\pazocal{H}}
\newcommand{\bR}{\mathbb{R}}
\newcommand{\bP}{\mathbb{P}}
\newcommand{\bp}{\mathbb{p}}
\newcommand{\hP}{\widehat{P}}
\newcommand{\hp}{\widehat{p}}

\newcommand{\up}{\breve{{p}}}
\newcommand{\bPu}{\mathbb{P}_{u,v}}
\newcommand{\bE}{\mathbb{E}}
\newcommand{\e}{\mathbf{e}}

\newcommand{\Ps}{{P}^\star}
\newcommand{\pp}{{\phi, \psi}}

\newcommand{\logsumexp}{\mathrm{log~sum~exp}}

\newcommand{\cred}{\color{black}}
\begin{document}
\title{Safety-Focused Reinforcement Learning Using Risk-Sensitive Coupled Control}
\author{Jiabin Lin, Shana Moothedath, \IEEEmembership{Senior Member, IEEE}, and Vivek S. Borkar \IEEEmembership{Fellow, IEEE}
\thanks{J. Lin and S. Moothedath are with Electrical and Computer Engineering, Iowa State University. Email: \{jiabin, mshana\}@iastate.edu. V. S. Borkar is with Department of Electrical Engineering, Indian Institute of Technology, Bombay, India. Email: borkar.vs@gmail.com}
}
\maketitle
\thispagestyle{headings}

\begin{abstract}
This paper investigates the coupled control problem within safety-focused reinforcement learning, addressing the essential challenge of achieving optimal performance while ensuring safety in systems operating within complex environments.
 We define safety as the requirement to prevent the system from reaching undesirable or hazardous states.
We propose a coupled control approach that prioritizes safety by employing effective methods to prevent the system from entering unsafe states while optimizing the performance metric, defined as the average cost.
We propose a novel problem transformation that reformulates it into a risk-sensitive problem combined with an average reward criterion.
We present a two-timescale Q-learning algorithm where in the faster timescale, the algorithm solves a risk-sensitive control problem to ensure safety, while on a slower timescale, the algorithm focuses on average cost control. 
We provide theoretical guarantees for the convergence of our algorithms, establishing their significance in ensuring safe, optimal control in dynamic environments. 
We validated the performance of our approach using GridWorld and CarGoal task in
Safety Gymnasium.
\end{abstract}
\begin{IEEEkeywords}
Safe control, constrained reinforcement learning, Q-learning, stochastic approximation
\end{IEEEkeywords}
\section{Introduction}
Markov decision processes (MDPs) are extensively used to model decision-making challenges in domains like autonomous vehicles  and robotics \cite{altman1999constrained}.
  As the focus on safety-critical cyber-physical systems intensifies, there is an increasing need to embed safety constraints within MDP models to balance performance with operational safety.  
 In applications such as autonomous vehicles, robotics, and industrial automation, integrating safety into the control framework is essential for reliable operation in dynamic, unpredictable environments. Coupled control enables systems to maintain safety in the presence of obstacles, failures, and disturbances while optimizing performance.
This paper presents the Safety-focused Coupled-Control (SCC) problem, aimed at minimizing cumulative costs while optimizing performance criteria and ensuring that system states remain within a defined safe set. By dividing the control strategy into two components, emergency control for safety and regular control for performance optimization, we create a structured approach to decision-making in high-stakes environments. This integration effectively addresses safety constraints while paving the way for more advanced hierarchical decision-making architectures.

Safe reinforcement learning (RL) has emerged as a crucial area of research, focusing on ensuring that learning agents operate within defined safety constraints while interacting with complex environments. 
For a survey on safe RL we refer to \cite{gu2022review, garcia2015comprehensive}.
A standard and well-studied formulation for reinforcement learning with constraints is the constrained Markov Decision Process (CMDP) framework \cite{borkar2005actor, altman2021constrained}, where agents must satisfy constraints on expectations of auxiliary costs \cite{hasanzadezonuzy2021learning, sui2015safe, achiam2017constrained}.
CMDP problems are well studied,  asymptotic \cite{altman1999constrained, bhatnagar2012online,paternain2019constrained,achiam2017constrained, tessler2018reward}, and non- asymptotic  \cite{chaudhary2022safe, liu2021learning, bura2022dope, ding2020natural, liu2021fast, zheng2020constrained} guarantees are provided.
The safety specifications in the majority of the CMDP methods are expected discounted cumulative costs \cite{altman1999constrained, achiam2017constrained, tessler2018reward, yang2020projection}. However, the main deficiency of expected costs is averaging the potential danger at a state to the whole trajectory.
For instance, an autonomous vehicle must maintain a safe distance from other traffic participants and obstacles at all times, not just an average or cumulative distance over a period, as a collision can occur in a single time step. Therefore, constraints based on expected cumulative costs do not provide a rigorous definition or guarantee absolute safety.
Control theoretic approaches based on  Lyapunov functions \cite{choi2020reinforcement, chow2018lyapunov, berkenkamp2017safe} and control barrier functions \cite{cheng2019end, taylor2020learning,ames2019control} were explored recently for safe learning.
In the former, one defines a Lyapunov function ensuring that learned policies are both optimal in terms of performance and safety. In the latter, a barrier function is designed to return a high penalty when the agent is close to an unsafe state. 
However, these methods rely on the prior formulation of the Lyapunov function, which results in conservative feasible sets. 
Hamilton-Jacobi (HJ) reachability analysis was also adapted for safe RL \cite{fisac2019bridging, hsu2021safety}.
However, these studies focus on ensuring safety in policy development without considering the performance criterion, such as reward optimization.
This safety-only design significantly limits broader applications of HJ reachability analysis in RL.

A recent study \cite{yu2022reachability} examined the safe RL problem as a reachability-constrained RL problem, where the goal is to ensure the entire state trajectory is safe, which is related to our problem setting.
\cite{yu2022reachability} considered a deterministic environment and policy and a discounted reward model for the objective. When the environment and policy are both deterministic, there is a unique trajectory for starting from each state (i.e., the future state is uniquely determined), which simplifies the approach, unlike our setting.
Later \cite{ganai2024iterative} extended the reachability-constrained RL problem to the stochastic setting.
The goal in  \cite{yu2022reachability, ganai2024iterative} is to find a policy that maximizes performance in safe states while minimizing cumulative damage in unsafe states.
To this end, they modeled the problem as a constraint optimization using a weighted linear combination of the two objectives and proposed a Lagrange relaxation-based algorithm similar to the CMDP approaches. 
In contrast, our approach seeks an optimal policy that enhances performance while strictly preventing the system from reaching predefined unsafe states. 
Additionally, we adopt an average cost performance model, which is harder to analyze, so the approach in \cite{yu2022reachability, ganai2024iterative} for the discounted setting does not apply directly.

In risk-sensitive sequential decision-making, the objective is to maximize a risk-sensitive criterion
such as the expected exponential utility \cite{borkar2002q, borkar2001sensitivity, moharrami2024policy}, a variance-related measure \cite{tamar2012policy, la2013actor, patton2022distributional, chow2015risk}, or the percentile performance \cite{filar1995percentile}. 
Risk-sensitive decision-making is also used as an alternative approach for handling uncertainty, which intriguingly
shares an elegant equivalence with robust decision-making \cite{zhang2023regularized,osogami2012robustness}.
The challenge in risk-sensitive RL involves the non-linearity of the objective function and the need for a risk-aware exploration mechanism.
Classical RL algorithms are unsuitable in this setting, as their design crucially relies on the linearity of Bellman equations.
To address these challenges, model-free algorithms such as Risk-Sensitive Value Iteration (RSVI) and Risk-Sensitive Q-learning (RSQ) have been introduced.

In this paper, we study the safe RL under average cost criteria. 
The objective of the agent is to develop a control policy that guarantees the state trajectory avoids any unsafe states while also minimizing the average cost associated with performance during operation.
This is particularly relevant in safety-critical applications like autonomous driving, where it is essential to ensure that unsafe states are never encountered, as even a single misstep can have catastrophic consequences. In such settings, where long-term stability is key, average reward ensures that the agent does not focus on maximizing immediate rewards at the cost of potential long-term risks.

\noindent{\bf Contributions:} Our contributions are threefold.

\noindent {$\bullet$ \em Safety-focused Coupled-Control problem (SCC):} 
We formulate SCC, a state safety-constrained control problem. The control policy consists of two interacting components: an emergency controller, which enforces state safety, and a regular controller, which optimizes performance subject to the safety constraints.
To solve the SCC problem, we introduce a novel problem transformation that converts the safety objective into a risk-sensitive criterion. Simultaneously, the performance objective is reformulated as an average-cost problem on a redefined Markov chain with a transformed probability structure that inherently enforces safety.

\noindent {$\bullet$ \em Two-timescale reinforcement learning algorithm for SCC:} We propose a two-timescale Q-learning algorithm, Q-SCC, to address the coupled-control problem. First, we transform the emergency control problem into a risk-sensitive formulation using a carefully defined problem transformation and introduce a Q-learning algorithm to solve this component.
The key challenge is that, unlike in standard risk-sensitive learning, where the risk cost is predefined and observed, in our setting the risk cost is induced by the safety requirements and hence unknown. To overcome this, we propose an eigenpair estimate of the kernel.
For the regular control problem, we develop a Q-learning algorithm under an average reward criterion, applying a kernel transformation to create a twisted kernel that respects the safety constraints. On a faster timescale, the Q-learning algorithm handles the risk-sensitive control, while on a slower timescale, the Q-learning algorithm focuses on average cost control.  We provide theoretical guarantees for the convergence of our algorithm, ensuring its effectiveness in managing both state safety and average reward performance objectives.

\noindent {$\bullet$ \em Numerical evaluation:} We evaluated our proposed Q-SCC algorithm on a GridWorld and a fixed-layout CarGoal task in Safety Gymnasium, and compared against benchmark methods, validating the effectiveness and reliability of our coupled control framework in achieving safety and performance. 
\section{Background and Related Work}
\subsection{Background: Markov Decision Process (MDP)}
MDP  is the standard framework for modeling a
stochastic dynamical system and computing its optimal control policy \cite{sutton2018reinforcement}. 
 Let $\X$ and $\U$ be compact sets describing the states and actions of the controller (agent), respectively, and $c:\X \times \U \rightarrow \bR$ is the cost function.  The system dynamics is characterized by the probability transition structure $\bP$, where $\bP(j|i, u)$ is the probability of transitioning to state $j$ from state $i$ under control action $u$. A  policy $\pi: \X \rightarrow \U$ is a conditional distribution $\pi(u|x)$ that guides the decision-making process of an agent. At time $t \in \{1,2,\ldots, n\}$, the agent observes the current state $X_t$ and chooses an action $U_t$ from the policy $\pi(u|i)$, and observes the cost $c(i, u)$.  The action chosen by the agent at a state drives the agent to the next state according to $\bP(j|i, u)$, where $j$ belongs to the set of neighboring states of $i$. A trajectory $\tau = \{X_0, U_0, X_1, U_1, \ldots, X_{n-1}, U_{n-1} \}$ is a sequence of state-action pairs traversed by the agent following a policy.
 The goal of the agent is to find an optimal policy $\pi^\star$ that minimizes the average cumulative cost $\lim_{n \rightarrow \infty}\frac{1}{n} \sum_{t=0}^{n-1} \bE [c(X_t, U_t)]$ under average cost criteria. %
  When the model $\bP$ is known, $\pi^\star$ can be computed using dynamic programming approaches, such as value iteration and policy iteration \cite{sutton2018reinforcement}.
  Here, the value function $V: \X \rightarrow \bR$ gives the solution to 
  \[V(i) =\min_{u \in \U}Q(i, u)=\min_{u\in \U}\big[\sum_{j}\bp(j|i,u)(c(i,u)+V(j))-\beta\big],\]
  where $\beta$ is the optimal cost. 
   $V(i)=\min_{u \in \U}Q(i, u).$
   In {\cred relative Q-value iteration}, Q-values are updated as
   \begin{equation*}\label{eq:Q-basic}
       Q_{n+1}(i, u)=\sum_{j}\bp(j|i,u)\big(c(i,u)+ \min_{u'\in \U} Q_n(j, u')\big)-Q_n(i_0, u_0)
   \end{equation*}
with $i_0\in \X$ and $u_0\in \U$ prescribed.
However, in many real-world problems, the model ($\bP$) is unknown. 
The goal of reinforcement learning algorithms is to learn the optimal policy without prior knowledge of the model.
\subsection{Related Work}
Constrained reinforcement learning (CRL) problems are typically modeled as Constrained Markov Decision Processes (CMDPs) \cite{altman2021constrained}. This approach extends traditional MDPs by incorporating a cost function to ensure compliance with constraints. The goal is to maximize the expected performance while keeping the cost incurred below a specified threshold.
Constrained optimization approaches are adopted to solve CRL problems. In penalty methods, penalty terms are introduced in the reward function to add safety constraints, discouraging the agent from taking actions that violate the safety criteria.
Lagrangian methods \cite{tessler2018reward, chow2017risk, ma2022joint,yu2022reachability, ganai2024iterative} incorporate safety constraints into the optimization problem by introducing Lagrange multipliers, which penalize constraint violations and balance the trade-off between performance and safety.
Trust-region methods are a well-studied approach \cite{achiam2017constrained, yang2020projection} that update the policy within a predefined safe region to prevent large, unsafe changes. These methods typically use Constrained Policy Optimization (CPO) with surrogate functions for rewards and safety constraints, ensuring that each policy update remains within a specific region. The process includes projection steps and backtracking line searches to iteratively adjust the policy while minimizing the risk of significant modifications.
Inspired by Proximal Policy Optimization (PPO) \cite{schulman2017proximal}, which built upon trust region policy optimization, the penalized proximal policy optimization approach \cite{zhang2022penalized} was introduced for safe reinforcement learning, enhancing trust-region methods by integrating a clipping mechanism to better balance exploration and safety. Constraint-Rectified Policy Optimization (CRPO) \cite{xu2021crpo} dynamically improves reward performance when constraints are satisfied, while actively minimizing constraint violations when they are not.
A model-based safe RL algorithm was proposed in \cite{bura2022dope} that integrates optimistic exploration for reward maximization with pessimistic exploration to prevent safety violations. This approach seeks to balance the exploration of high-reward actions while avoiding unsafe states. It relies on access to a known safety policy and combines a reward bonus for optimistic exploration with a conservative constraint for pessimistic exploration, in addition to the standard optimistic model-based exploration techniques.
Although all these algorithms offer benefits through various learning mechanisms, they still inherit the challenges of the CMDP framework. Specifically, they do not ensure absolute safety and are generally unsuitable for strict state-wise safety.
Recently, safe RL has been studied in the context of reachability-constrained RL problem \cite{yu2022reachability, ganai2024iterative}. 
Reachability-constrained RL uses reachability to optimize and maintain persistent safety in the feasible set via a cumulative cost-minimization objective along with discounted reward criteria for the performance. 
Both works leveraged Lagrangian method to solve the problem and provided convergence guarantees under certain assumptions on differentiability and Lipchitzness.

\section{Problem Formulation}

We consider a controlled stochastic matrix $\bPu(j|i)$, where $i,j$ belong to a finite state space $\X$ and $u,v$ are control variables taking values in finite action spaces $\U, \V$, respectively.
We denote $\phi(\cdot), \psi(\cdot)$ respectively, a generic stationary or stationary randomized policy. We use $\{X_n\}$ to denote a generic controlled Markov chain and $\{U_n\}, \{V_n\}$ the corresponding $\U-$valued, $\V-$valued control sequences, respectively.

Our goal is to model the coupled-control problem, where the goal is to devise a joint control policy that minimizes the expected cost for a performance criterion while ensuring that the state of the system remains within the safe set $\X_0 \subset \X$. We refer to the control $\{U_n\}$ as the {\bf regular control} and $\{V_n\}$ as the {\bf safety/emergency control}. 
The regular control is trying to minimize a given average cost, and the emergency control is trying to keep the state within $\X_0$.
{\cred The two controls represent separate task and safety measures. A single controller over the joint action $(u, v)$ could represent the same policy. Our aim is to impose distinct optimization requirements. The emergency action is evaluated against the worst regular action through the max–min objective. Thus, its protective role does not rely on the regular controller already choosing for its most efficient action. The safe set provides the safe observed states but does not indicate which actions maintain the safety of the next state when the transition probabilities are unknown.
Mathematically, the regular-control problem is formulated as}
\begin{equation}\label{eq:regular_cost}
    \lim_{n \rightarrow \infty} \frac{1}{n}\sum_{t=0}^{n-1}\bE\Big[w(X_t, U_t)\Big],
\end{equation}
where $w(\cdot)$ is the running cost for each state-control (regular). 

The goal of the emergency control is to keep the state within $\X_0$. 
Let $\bPu^0$ denote the restriction of $\bPu$ to $\X_0$, {\cred and $\bP^0_{\phi, \psi}$ the corresponding matrix under a fixed stationary policy pair $(\phi, \psi)$}. Without loss of generality, we assume $\X_0=\{1, 2, \ldots, |\X_0|\}$. Then {\cred $\bP^0_{\phi, \psi} \in \bR^{|\X_0| \times |\X_0|}$} is substochastic and has a unique principal eigenvalue {\cred $\Lambda \leqslant 1$} and a component-wise strictly positive principal eigenvector $\e=[e(1), \ldots, e(|\X_0|)]$, rendered unique by a suitable normalization. Let us denote $H:=\min\{n \geq 0:X_n \notin \X_0\}$ for a controlled Markov chain $\{X_n\}$ {\cred under this policy pair}. 
Using $\bf{1}$ to denote the $|\X_0|-$dimensional vector of all ones, we have
\begin{equation*}
\mathrm{Prob}(H > n) = \mu {\cred (\bP^0_{\phi, \psi})^n} \bf{1},
\end{equation*}
where $\mu \in \bR^{1 \times |\X_0|}$ is the initial distribution on $\X_0$. Thus,
\begin{equation}\label{eq: path}
\lambda := \log \Lambda = \lim_{n \rightarrow \infty}\frac{1}{n}\log \Big(\mu {\cred (\bP^0_{\phi, \psi})^n} \bf{1}\Big).
\end{equation}
In Eq.~\eqref{eq: path}, the right-hand side is simply the logarithm of the probability of all trajectories of length $n$ that never leave $\X_0$. Thus, the safety/emergency control problem aims at maximizing $\lambda$.
We assume that there exists at least one admissible $\{V_n\}$ that meets the safety requirement.

The overall objective of the Safety-focused Coupled-Control problem (SCC) problem is to minimize the regular average cost in Eq.~\eqref{eq:regular_cost} while maximizing the emergency safety exponent in Eq.~\eqref{eq: path}.
We make the following standard assumptions.

\begin{assu}\label{assume:irr}
We assume that the restricted state space $\X_0$ is irreducible and aperiodic for any choice of $(\phi, \psi)$.
\end{assu}
Assumption~\ref{assume:irr} is standard in reinforcement learning and is crucial for the convergence of learning algorithms, allowing for accurate estimation of optimal policies and value functions across diverse and complex environments \cite{sutton2018reinforcement, puterman2014markov}.
\section{Proposed Solution: Two-Timescale Q-Learning Algorithm for SCC}
First, we reformulate the safety-constrained (emergency) problem as outlined below. This approach leverages the insight that, with appropriate definitions, the emergency problem can be framed as a risk-sensitive optimization problem.
We define $\Ps_{u,v}:= \big[p^\star(j|i, u, v) \big]_{i,j \in \X_0}$ be a stochastic matrix defined as
\[p^\star(j|i, u, v):=\frac{\bp(j|i, u, v)}{d_{u,v}(i)}, \mbox{~where~} d_{u,v}(i)= \sum_{j \in \X_0} \bp(j|i, u, v).\]
We define the $|\X_0| \times |\X_0|$ diagonal matrix $D_{u,v}$ with the $i-$th diagonal element as $d_{u,v}(i)$ for all $i \in \X_0$. 
Let $c_{u,v}(i):= \log d_{u, v}(i)$, for all $i \in \X_0$.
By direct verification, we get
\[\bP^0_{u,v}=D_{u, v}P^\star_{u, v},\]
with $D_{u, v}(i)=e^{c_{u,v}(i)}.$ 
We assume the well-posed condition $d_{u,v}(i) > 0$ for all $(i, u, v) \in \X_0 \times \U \times \V$. This ensures that the normalized kernel $P_{u,v}^\star$ and the logarithmic cost $c_{u,v}(i) = \log d_{u,v}(i)$ are well defined for every state-action pair.
Thus the emergency control problem in Eq.~\eqref{eq: path} is can be written as
\begin{eqnarray}
    &\lambda = \log \Lambda = \lim_{n \rightarrow \infty}\frac{1}{n}\log \big(\mu {\cred (\bP^0_{\phi, \psi})^n} {\bf{1}}\big)\nn\\
    &=\lim_{n \rightarrow \infty}\frac{1}{n}\log \bE\big[e^{\sum_{t=0}^{n-1}c_{U_t, V_t}(X_t)} \big]\label{eq: path_new}
\end{eqnarray}
for an $\X_0-$valued controlled Markov chain with controlled transition matrix $\Ps_{u,v}$ {\cred under the same policy pair}.
Since emergency control is responsible for ensuring safety, we adopt a risk-sensitive safety criterion
$$
\lambda^\star := \sup_{\psi} \inf_{\phi} \lim_{n \rightarrow \infty} \frac{1}{n} \log \bE^{\phi, \psi} \big[\exp \big(\sum_{t=0}^{n-1} c_{U_t, V_t}(X_t)\big)\big],
$$
where $\phi$ ranges over admissible regular-control policies and $\psi$ ranges over admissible emergency-control policies. Thus, under the above problem transformation, the emergency control problem is reformulated as the risk-sensitive control problem
\begin{equation}\label{eq: emerg}
\psi^\star\in\arg\sup_{\psi}\inf_{\phi}\lim_{n\rightarrow\infty}\frac{1}{n}\log \bE^{\phi,\psi}\big[\exp\big(\sum_{t=0}^{n-1} c_{U_t, V_t}(X_t)\big)\big].
\end{equation}
Equivalently, at the dynamic-programming level, the Bellman-Perron operator has the structure $\max_{v \in \V} \min_{u \in \U}$. 

Next, we discuss the regular control problem. Consider stationary randomized policies, $\phi, \psi$.
Let us denote the principal eigenvalue-eigenvector pair of the transition matrix {\cred $\bP_{\phi, \psi}^0$} under policy $\phi, \psi$ as $(\Lambda_\pp, \e_\pp)$. We define $|\X_0| \times |\X_0| $ matrix
\begin{equation}\label{eq:twist}\hP_{\pp} := \big[ \hp_\pp(j|i) \big]_{i,j \in \X_0},~\hp_\pp(j|i):= \frac{{\cred p_\pp(j|i)} \e_{\phi, \psi}(j)}{\Lambda_\pp \e_{\phi, \psi}(i)}.\end{equation}
From Theorem~1 \cite{avrachenkov2010quasi} and Assumption~\ref{assume:irr}, $\hP_{\pp}$ is a transition probability matrix for a Markov chain governed by the policy pair $\pp$ {\cred conditioned on remaining in the safe state set $\X_0$ up to a horizon tending to infinity}.
{\cred The eigenvector ratio weights safe successor states according to their long-term survival under the fixed policy pair.}
{\cred Hence, for the regular-control problem solved by Q-SCC, the expectation in Eq.~\eqref{eq:regular_prob} uses the limiting conditional mean of Eq.~\eqref{eq:twist-2}, with the limiting emergency-action distribution held fixed as in Section~\ref{Convergence_Regular}. This kernel uses the fixed-action eigenpairs in Eqs. ~\eqref{principal_eigenvector}–~\eqref{principal_eigenvalue}.}
Mathematically, we can rewrite the regular control problem as, under $\hP_{\pp}$
\begin{equation}\label{eq:regular_prob}
\phi^\star\in \arg\min_\phi  \Big(  \lim_{n \rightarrow \infty} \frac{1}{n}\sum_{t=0}^{n-1}\bE^{\phi}\Big[w(X_t, U_t)\Big]\Big),
\end{equation}
where $w(\cdot)$ is a given running cost function defined for a state-control (regular) action pair. The couple-control SCC problem hence is to solve for $(\phi^\star, \psi^\star)$ given in Eq~.\eqref{eq: emerg} and Eq.~\eqref{eq:regular_prob}.
\subsection{Q-SCC: Two-Timescale Q-learning Algorithm for Safety-Focused Coupled-Control}
	\begin{algorithm}[!htbp]
		\caption{Q-SCC: Two-Timescale Q-learning for SCC}
		\label{algo:Q}
		\begin{algorithmic}[1]
			\State \textbf{Input:} State space $\X$, safe set $\X_0$, action sets $\U$, $\V$, cost functions $w(\cdot)$, exploration parameter $\epsilon \in (0, 1)$, a margin parameter $\Delta > 0$, and number of episodes $T >> 0$
            \State \textbf{Output:} {\cred Learned policies $({\phi}^{T}, {\psi}^{T})$}
                \State Initialize randomly the emergency control Q-values $Q_0(i, u, v) > 0$ for $(i, u, v) \in \X_0 \times \U \times \V$, and $Q_0(i, u, v) \leftarrow 0$ for $i \notin \X_0$
                \State Initialize randomly principal eigenvector $\e_0(i, u, v) > 0$ for $(i, u, v) \in \X_0 \times \U \times \V$, and $\e_0(i, u, v) \leftarrow 0$ for $i \notin \X_0$
                \State Initialize randomly the regular control Q-values $\widetilde{Q}_0(i,u)$, for $i \in \X_0$, and $\widetilde{Q}_0(i,u) \leftarrow 0$, for $i \notin \X_0$, for all $u \in \U$
                \State Initialize the global counter $n \leftarrow 0$ and the visit counters $\nu(i, u, v, 0) \leftarrow 0$ for all $(i, u, v) \in \X_0 \times \U \times \V$. Fix a deterministic ordering on $\V$ for tie-breaking and choose an initial emergency action $\bar{v}_0(i) \in \V$ for each $i \in \X_0$.
            \For{episode $m = 1, 2, \cdots, T$}
                \State Initialize $t \leftarrow 0$, $X_0^{(m)} \in \X_0$
                \While  {$X_t^{(m)} \in \X_0$}
                    \State 
                    For all $k \neq X_t^{(m)}$, set $\bar{v}_{n+1}(k) = \bar{v}_n(k)$. For the current state $X_t^{(m)}$, update the emergency action by
                    \begin{equation*}
                    \bar{v}_{n+1}(X_t^{(m)}) = 
                    \left\{
                    \begin{aligned}
    			    &\bar{v}_n(X_t^{(m)}), \text{if }
                    \max_{v' \in \V} \min_{u' \in \U} Q_n(X_t^{(m)}, u', v')\\
                    &\leqslant \min_{u' \in \U} Q_n(X_t^{(m)}, u', \bar{v}_n(X_t^{(m)})) + \Delta \\
                    &\text{the first element of}\\
                    &\arg\max_{v' \in \V} \min_{u' \in \U} Q_n(X_t^{(m)}, u',v'), \text{else}.
                    \end{aligned}
    			    \right.
                    \end{equation*} 
                    \State $V_n = 
                    \begin{cases}
    			    \bar{v}_{n+1}(X_t^{(m)}), & \mbox{~w.p.~} 1 - \epsilon \\
                    \mbox{~uniformly~sample~} v \sim \V, & \mbox{~w.p.~} \epsilon.
    			    \end{cases}$
                    \State $U_n \in \begin{cases}
			        \arg\min_{u' \in \U} \widetilde{Q}_n (X_t^{(m)}, u'), & \mbox{~w.p.~} 1-\epsilon\\
                    \mbox{~uniformly~sample~} u\sim \U, & \mbox{~w.p.~} \epsilon.
			        \end{cases}$
                    \State Apply control $U_n, V_n$ at $X_t^{(m)}$ and observe cost $w(X_t^{(m)}, U_n)$ and next state $X_{t+1}^{(m)}$
                    \State Update emergency Q iterates as in Eq.~\eqref{eq:Q}
                    \State Update principal eigenpair iterates as Eqs.~\eqref{principal_eigenvector}, \eqref{principal_eigenvalue}
                    \State Update the regular control Q iterates as in Eq.~\eqref{eq:Q-regular}
                    \State Update global visit counter $\nu(X_t^{(m)}, U_n, V_n, n)$
                    \State Set $n \leftarrow n+1$ and $t \leftarrow t+1$
                \EndWhile
            \EndFor
		\end{algorithmic}\label{algo}
	\end{algorithm}

Now we present the proposed algorithm to solve the coupled-control problem. The pseudocode is given in Algorithm~\ref{algo:Q}. Our approach is a two-timescale RL algorithm, comprising a fast-timescale safety update and a slow-timescale performance update. The fast update consists of two coupled stochastic-approximation recursions. We apply a normalized risk-sensitive Q-learning recursion, inspired by Borkar's relative Q-learning framework for risk-sensitive control \cite{borkar2002q}, to address the safety/emergency problem. 
The key challenge is that, unlike in standard risk-sensitive Q-learning, where the immediate cost is observed and known, in our setting the cost is induced by the state safety constraint. Consequently, the cost model depends on the transition dynamics, which are unknown a priori. We therefore perform an eigen estimate.
We then analyze regular control as a worst-case scenario with respect to the safety problem. Thus, the corresponding Bellman-Perron operator provides an effective structure $\max_{v \in V} \min_{u \in U}$, resulting in the emergency update presented in Eq.~\eqref{eq:Q}. 

Algorithm~\ref{algo:Q} maintains a state-dependent emergency selector $\bar{v}_n: \X_0 \rightarrow \V$. Upon visiting state $i$, the margin-inertia rule sets  $\bar{v}_{n+1}(i) = \bar{v}_n(i)$ if $\max_{v \in \V} \min_{u \in \U} Q_n(i, u, v) \leqslant \min_{u \in \U} Q_n(i, u, \bar{v}_n(i)) + \Delta$. Otherwise, $\bar{v}_{n+1}(i)$ is assigned the first maximizing action according to the specified ordering of $\V$. Thus, a switch occurs only when the alternative improves the current $Q$-value by more than a threshold $\Delta$, thus preventing repeated switching among very similar emergency actions. The rule maintains the Bellman–Perron objective in Eq.~\eqref{eq:Q} unaffected, with the optimization $\max_{v \in \V} \min_{u \in \U}$.

{\cred The regular-control problem is formulated using the limiting kernel associated with Eq.~\eqref{eq:twist-2}.}
The related twisted kernel requires the principal eigenvalue-eigenvector pair of the substochastic safe kernel. Given the unknown nature of this pair, Q-SCC applies an asynchronous stochastic approximation recursion to estimate the principal eigenpair, as shown in Eqs.~\eqref{principal_eigenvector} and \eqref{principal_eigenvalue}. This eigenpair estimation step is significantly different from the emergency risk-sensitive Q-update in Eq.~\eqref{eq:Q}. Eq.~\eqref{eq:Q} determines the safety action-value fixed point, while Eqs.~\eqref{principal_eigenvector} and \eqref{principal_eigenvalue} obtain the scalar values needed to develop the twisted kernel. 

In the slower timescale, driven by $b(n)$ where $\frac{b(n)}{a(n)} \rightarrow 0$, the regular controller applies relative Q-learning for average cost MDPs. Thus, the slow regular control recursion treats the emergency updates and the scalar values derived by the fast stage as quasi-static. Thus, the emergency-control Q-learning updates using stochastic approximation is $Q_{n+1}(i,u,v)=$
\begin{eqnarray}
& Q_{n}(i,u, v) + \mathbb{1}_{\{X_n=i, U_n=u, V_n=v\}} a(\nu(i,u,v,n)) \nonumber\\
\hspace{-17 mm}&\Big(\mathbb{1}_{\{X_{n+1}\in \X_0\}}\frac{\max\limits_{v'\in \V}\min\limits_{u'\in\U}Q_{n}(X_{n+1}, u', v')}{Q_n(i_0,u_0,v_0)}-Q_{n}(i,u,v) \Big)\label{eq:Q},
\end{eqnarray}
where $\mathbb{1}_{\{f\}}$ is an indicator function with $\mathbb{1}_{\{f\}}=1$ if $f$ holds and $\mathbb{1}_{\{f\}}=0$, otherwise, and the state action pair $i_0$ and $u_0, v_0$ are fixed elements of $\X_0, \U, \V$, respectively.
The notation $\nu(i,u,v,n)$ here denotes the number of times the state-action pair $(i, u,v)$ is updated until $n$.
\begin{assu}\label{Viable_Safe}
There exist constants $C_H < \infty$, $c_H > 0$, and a finite random time
$N_H < \infty$ such that, for all $n \geqslant N_H$ and all $(i, u, v) \in \X_0 \times \U \times \V$, almost surely
\begin{eqnarray*}
\bE\Big[\Big(\mathbb{1}_{\{X_{n+1} \in \X_0\}} \frac{\max_{v' \in \V} \min_{u' \in \U} Q_{n}(X_{n+1}, u', v')}{Q_n(i_0, u_0, v_0)}\Big)^2|\\
\pazocal{F}_n, X_n = i, U_n = u, V_n = v \Big] \leqslant C_H^2,
\end{eqnarray*}
and the anchor component satisfies
\begin{eqnarray*}
\bE\Big[\mathbb{1}_{\{X_{n+1} \in \X_0\}} \frac{\max_{v' \in \V} \min_{u' \in \U} Q_{n}(X_{n+1}, u', v')}{Q_n(i_0, u_0, v_0)} |\\
\pazocal{F}_n, X_n = i_0, U_n = u_0, V_n = v_0\Big] \geqslant c_H.
\end{eqnarray*}
\end{assu}
Also, stepsize $\{a(n)\}$ satisfy the Robbins-Monro conditions
\begin{eqnarray*}
&\nu(i, u, v, n) = \sum_{t=0}^n \mathbb{1}_{\{X_t=i, U_t=u, V_t=v\}}, \quad n \geqslant 0,\\
&0 < a(n) < 1, \quad \sum_{n}a(n) = \infty, \mbox{~and~} \sum_{t}a(n)^2<\infty.
\end{eqnarray*}
Let $A(n) := \sum_{\ell=0}^{n} a(\ell)$. For every $x \in (0, 1]$,
\begin{equation*}
\lim_{n \rightarrow \infty} \frac{A(\lfloor xn \rfloor)}{A(n)} \rightarrow 1, \quad \sup_{n \geqslant 1} \frac{a(\lfloor xn \rfloor)}{a(n)} < \infty.
\end{equation*}
Furthermore, every component $(i, u, v) \in \X_0 \times \U \times \V$ is sampled infinitely often and there exists $\eta > 0$ such that
\begin{equation*}
\liminf_{n \to \infty} \frac{\nu(i, u, v, n)}{n} \geqslant \eta, \quad \text{~for~all~} (i, u, v) \in \X_0 \times \U \times \V.
\end{equation*}
Eq.~\eqref{eq:Q} is the model-free sample variant of the normalized Bellman-Perron update related to the risk sensitive emergency problem. Given that $X_n = i$, $U_n = u$, and $V_n = v$,
\begin{align*}
&\bE\Big[\mathbb{1}_{\{X_{n+1} \in \X_0\}} \max_{v' \in \V} \min_{u' \in \U} Q_n(X_{n+1}, u', v') |X_n, U_n, V_n\Big] \\
&= e^{c_{u, v}(i)} \sum_{j \in \X_0} P_{u, v}^\star (j | i) \max_{v' \in \V} \min_{u' \in \U} Q_n(j, u', v').
\end{align*}
Thus, the indicator $\mathbb{1}_{\{X_{n+1} \in \X_0\}}$ provides an unbiased sample of the substochastic safe transition operator. In particular, the emergency update in Eq.~\eqref{eq:Q} does not require a specific estimation of $d_{u, v}(i) = e^{c_{u,v}(i)}$. The normalization via the anchor component $Q_n(i_0, u_0, v_0)$ is the relative normalization derived from risk sensitive Q-learning, which determines the scale of the positive fixed point. Under the normalization $Q^\star(i_0, u_0, v_0) = \Lambda^\star$, the limiting mean field of Eq.~\eqref{eq:Q} is the normalized ODE $\dot{Q} = G(Q) - Q.$
For the regular control, we utilize the relative value iteration Q-learning algorithm from \cite{abounadi2001learning} with the
empirical row-wise twisted kernel in Eq.~\eqref{eq:twist-2}. The corresponding relative Bellman target for $(i, u)$ 
\begin{equation*}
w(i,u) + \sum_{j \in \X_0} \breve{p}_n(j | i, u) \min_{u' \in \U} \widetilde{Q}_n(j, u') - \min_{u' \in \U} \widetilde{Q}_n(i_0, u'),
\end{equation*}
where $i_0 \in \X_0$ is fixed and
{\cred \begin{eqnarray}\label{eq:twist-2}
\up_n(j|i, u) := \frac{\mathbb{1}_{\{j \in \X_0\}} \bP(j|i, u, V_n) \e_n(j, u, V_n)}{\Lambda_n(u, V_n) \e_n(i, u, V_n)}. \label{twisted_p}
\end{eqnarray}}
The construction of the twisted kernel requires the principal eigenpair of the substochastic safe kernel related to the current action pair. Due to the unknown transition model, these scalar values must also be learned from samples. Thus, we perform an asynchronous Perron-Frobenius eigenpair recursion in a fast timescale, drawing inspiration from the stochastic approximation spectral-ranking approach \cite{borkar2014asynchronous}. For each fixed pair $(u, v)$, the vector $e_n(\cdot, u, v)$ estimates the positive principal eigenvector of the corresponding safe kernel, while $\Lambda_n(u, v)$ estimates its principal eigenvalue under the normalization given by the positive weight vector $\alpha$.
For prescribed weights $\alpha = [\alpha_1, \cdots, \alpha_{|\X_0|}]^{\top}$ with all $\alpha_i > 0$, 
\begin{eqnarray}
&\e_{n + 1}(i, u, v) = \e_n(i, u, v) + \mathbb{1}_{\{X_n=i, U_n=u, V_n=v\}} \nn\\
&a(\nu(i,u,v,n))\Big(\mathbb{1}_{\{X_{n+1}\in \X_0\}} \frac{\e_n(X_{n + 1}, u, v)}{\alpha^{\top} e_n(\cdot, u, v)} - \e_n(i, u, v)\Big) \label{principal_eigenvector}\\
&\Lambda_{n + 1}(u, v) = \alpha^\top \e_{n + 1}(\cdot, u, v). \label{principal_eigenvalue}
\end{eqnarray}

We now discuss the slower timescale regular control recursion. Based on the current emergency action $V_n$ chosen by the fast safety controller and the current principal eigenpair estimates $(\Lambda_n, e_n)$, we construct the estimated twisted kernel $\up_n(\cdot | i, u)$ in Eq.~\eqref{twisted_p}. This is the sample-based transition kernel used by the regular recursion. Upon the convergence of the fast variables, its limiting counterpart is a valid stochastic kernel on $\X_0$. The regular controller then performs the relative Q-learning method for average cost MDPs \cite{abounadi2001learning} with respect to this limiting twisted kernel. Given that $\frac{b(n)}{a(n)} \rightarrow 0$, the principal eigenpair estimates and the emergency selector proceed on the faster timescale are treated as quasi-static by the regular control recursion.
Thus the Q-update for regular control is
\begin{align}
&\widetilde{Q}_{n+1}(i,u) = \widetilde{Q}_{n}(i,u) + b(\nu_R(i, u, n)) \mathbb{1}_{\{X_n=i, U_n=u\}}\nn\\
&\big(w(i,u) + \mathbb{1}_{\{X_{n+1}\in \X_0\}} \times\frac{\e_{n + 1}(X_{n + 1}, u, V_n)}{\Lambda_{n + 1}(u, V_n) \e_{n + 1}(i, u, V_n)}\nn\\
&\min_{u' \in \U} \widetilde{Q}_n(X_{n+1}, u')- \min_{u' \in \U} \widetilde{Q}_n(i_0, u') - \widetilde{Q}_n(i, u)\big) \label{eq:Q-regular}
\end{align}
Here, $\nu_R(i, u, n) = \sum_{t=0}^{n} \mathbb{1}_{\{X_t = i, U_t = u\}}$ represents the regular-control counter. The sequence $b(n)$ satisfies the Robbins-Monro conditions, i.e., $\sum_{n}b(n)= \infty, \mbox{~and~} \sum_{t}b(n)^2<\infty.$
Further, we set $b(n) = o(a(n))$.
Let $B(n) := \sum_{\ell=0}^{n} b(\ell)$. For every $x \in (0, 1]$,
\begin{equation*}
\lim_{n \rightarrow \infty} \frac{B(\lfloor xn \rfloor)}{B(n)} \rightarrow 1, \quad \sup_{n \geqslant 1} \frac{b(\lfloor xn \rfloor)}{b(n)} < \infty.
\end{equation*}
Moreover, the corresponding regular-control sampling condition is derived from Assumption~\ref{Viable_Safe}. Indeed, since $\nu_R(i, u, n) = \sum_{v \in \V} \nu(i, u, v, n)$, Assumption~\ref{Viable_Safe} indicates that every regular component $(i, u) \in \X_0 \times \U$ is updated infinitely often. It follows 
$\liminf_{n \rightarrow \infty} \frac{\nu_R(i, u, n)}{n} \geqslant |\V|\eta$.

{\cred The Q-value and eigenpair tables require $O(|\X_0| |\U| |\V|)$ storage. A statewise max–min computation costs $O(|\U| |\V|)$, and direct eigenvector normalization costs $O(|\X_0|)$.}
\subsection{Convergence Analysis of Emergency Control}
In this subsection, we prove almost-sure convergence of the emergency-control recursion in Eq.~\eqref{eq:Q}. 
\begin{definition}\label{def:Th}
Define $T: \bR^{|\X_0|} \rightarrow \bR^{|\X_0|}$ by
\begin{equation*}
(Th)(i) := \max_{v}\min_{u}(c_{uv}(i)+\log\sum_{j\in\X_0}P_{uv}^{\star}(j|i)\exp(h(j))),
\end{equation*}
for $i \in \X_0$, and $\widehat{T}: \bR^{|\X_0|} \rightarrow \bR^{|\X_0|}$ by
\begin{eqnarray*}
&(\widehat{T}x)(i)=\lim_{a\rightarrow\infty}\frac{1}{a}(T(ax))(i)=\lim_{a\rightarrow\infty}\frac{1}{a}\max_{v}\min_{u}\\
&(c_{uv}(i)+\log\sum_{j\in\X_0}e^{ax(j)+\log{P_{uv}^{\star}(j|i)}}).
\end{eqnarray*}
Furthermore, define $Q$ by $Q(i,u,v) =$
\begin{equation*}
\frac{\exp(c_{uv}(i))}{\Lambda^\star} \sum_{j \in \X_0} P_{uv}^{\star}(j|i) \max_{v^{\prime}} \min_{u^{\prime}} Q(j,u^{\prime},v^{\prime}) \triangleq F_{iuv}(Q),
\end{equation*}
and for any positive function $q$ set 
\begin{equation}\label{eq:qset}
(q)_m = \min_{i,u,v} \frac{q(i,u,v)}{Q^{\star}(i,u,v)}, \quad (q)_M = \max_{i,u,v} \frac{q(i,u,v)}{Q^{\star}(i,u,v)},
\end{equation}
where $Q^\star$ denotes a strictly positive fixed point of $F$, i.e., $Q^\star = F(Q^\star)$. 
Moreover, define $G_{iuv}(Q)$ and $F_{iuv}(Q)$ by
\begin{eqnarray*}
G_{iuv}(Q) &= \frac{e^{c_{uv}(i)}}{Q(i_0,u_0,v_0)} \sum_{j\in\X_0} P_{uv}^{\star}(j|i) \max\limits_{v^{\prime}}\min\limits_{u^{\prime}} Q(j,u^{\prime},v^{\prime}), \\
F_{iuv}(Q) &= \frac{e^{c_{uv}(i)}}{\Lambda^{\star}} \sum_{j\in\X_0} P_{uv}^{\star}(j|i) \max\limits_{v^{\prime}}\min\limits_{u^{\prime}} Q(j,u^{\prime},v^{\prime}).
\end{eqnarray*}
\end{definition}
From the above definition, operator $T$ satisfies the following:
\begin{itemize}
    \item \emph{Translation invariance:} $T(h + \alpha \mathbf{1}) = Th + \alpha \mathbf{1}$ for all $\alpha \in \bR$, $h \in \bR^{|\X_0|}$, where $\mathbf{1}$ denotes the vector of all ones.
    \item \emph{Monotonicity:} if $h \leqslant g$ componentwise, then $Th \leqslant Tg$ componentwise.
    \item \emph{Nonexpansiveness:} $\|Th - Tg\|_{\infty} \leqslant \|h - g\|_{\infty}$ and $\|Th - Tg\|_{\H} \leqslant \|h - g\|_{\H}$, where $\|\cdot\|_\infty$ is the supermum-norm and $\|\cdot\|_{\H}$ denotes the Hilbert semi-norm. 
\end{itemize}
Thus, $G(Q) = \frac{\Lambda^\star}{Q(i_0, u_0, v_0)} F(Q)$. The representative of the positive fixed-point ray is normalized by $Q^\star(i_0, u_0, v_0) = \Lambda^\star$, resulting in $F(Q^\star) = G(Q^\star) = Q^\star$.

The analysis consists of three parts. 
In the first part, we prove the existence of the Bellman-Perron eigenpair, identify its eigenvalue with the optimal risk-sensitive safety exponent, and demonstrate that its positive eigenvector is unique up to a multiplicative constant. Thus, the given anchor normalization results in a unique fixed point $Q^\star$. In the second part, we investigate the limiting ODE and demonstrate that $Q^\star$ is its unique globally asymptotically stable equilibrium. 
Finally, in third part, we formulate the emergency-control update as an asynchronous stochastic-approximation recursion, ensure the required stability and martingale-noise conditions, and apply the ODE method to establish that $Q_n$ converges to $Q^\star$ almost surely. We present the main theorems and a proof sketch below. The complete proofs are in Appendix.
\begin{theorem}\label{eigen_relation}
Let $(\Lambda_1,H_1)$ and $(\Lambda_2,H_2)$ be two strictly positive solutions of $H = \frac{\exp(Th)}{\Lambda}$. Then $\Lambda_1 = \Lambda_2 = \Lambda^{\star}$ and $H_1 = c H_2$, where $\Lambda = \exp(\lambda)$ and $H = \exp(h)$.
\end{theorem}
\begin{theorem}\label{Risk:Thm3}
The vector $Q^{\star}$ is a globally asymptotically stable equilibrium point for the ODE $\dot{Q} = G(Q) - Q.$
\end{theorem}
\begin{theorem}\label{emergency_convergence}
Suppose Assumption~\ref{Viable_Safe} holds. Under Assumption~\ref{assume:irr}, the standing well-posedness condition $d_{u, v}(i) > 0$, the positive initialization in Algorithm~\ref{algo:Q}, and the sampling condition stated after Eq.~\eqref{eq:Q}, the emergency Q-learning recursion in Eq.~\eqref{eq:Q} satisfies $Q_n \rightarrow Q^\star \quad\text{almost surely}.$
Consequently, $Q_n(i_0, u_0, v_0) \rightarrow \Lambda^\star \quad\text{almost surely}.$
\end{theorem}
{\cred \begin{remark}
(Safety of the limiting policy) Given the assumptions of Theorem~\ref{emergency_convergence} and there exists a stationary emergency policy $\psi$ satisfying $d_{u, \psi(i)}(i) = 1$ for every $i \in \X_0$ and $u \in \U$, derive that $\max_v \min_u d_{u, v}(i) = 1$, leading to $T(0) = 0$. According to Theorem~\ref{eigen_relation}, it follows that $\Lambda^\star = 1$ and there exists a constant positive state eigenvector. Thus, Definition~\ref{def:Th} implies that $Q^\star(i, u, v)$ is proportional to $d_{u, v}(i)$, with the same positive factor for all $(i, u, v)$. Since $d_{u, v}(i) \leqslant 1$, every selector
\begin{equation*}
\psi^\star(i) \in \arg\max_{v \in \V} \min_{u \in \U} Q^\star(i, u, v)
\end{equation*}
satisfies $d_{u, \psi^\star(i)}(i) = 1$ for every regular action $u$. Performed without exploration, this policy keeps every transition within $\X_0$. Induction over finite horizons and continuity of probability yield
\begin{equation*}
\mathrm{Prob}(H = \infty | X_0 = i) = 1,
\quad i \in \X_0.
\end{equation*}
This guarantee concerns the exact limiting selector. It does not include exploratory training or policies obtained after an arbitrary finite training budget.
\end{remark}}
{\bf Proof Sketch:} First, Lemma~\ref{logsumexp} establishes the recession operator associated with the logarithmic Bellman-Perron operator. Lemma~\ref{fixed point} subsequently utilizes the irreducibility of the restricted safe kernels to demonstrate that this recession operator has only constant fixed points. The nonlinear Perron-Frobenius
argument in Theorem~\ref{Risk:Thm1} thus yields an additive eigenpair $Th = \lambda \mathbf{1} + h$. Lemma~\ref{lambda_equal} identifies $\lambda^\star$ with the optimal risk-sensitive safety exponent, while Theorem~\ref{eigen_relation} shows that $\Lambda^\star = \exp(\lambda^\star)$ is unique and that the positive eigenvector is unique up to multiplication by a positive scalar. Consequently, the given normalization gives a unique positive action-value fixed point $Q^\star$. Second, Lemma~\ref{support_lemma1_secondary} and Corollary~\ref{support_corollary1_secondary} compare a positive vector $q$ with $Q^\star$ through and demonstrate that $(q)_m \leqslant (F(q))_m \leqslant (F(q))_M \leqslant (q)_M$. As a result, along the unnormalized ODE $\dot{Q}(t) = F(Q(t)) - Q(t)$, the lower ratio shows nondecreasing behavior while the upper ratio demonstrates a nonincreasing trend (Lemma~\ref{support_lemma2_secondary}). Irreducibility eliminates the possibility of a nonzero limiting gap between these ratios, thus Theorem~\ref{Risk:Thm2} demonstrates that each positive solution approaches the ray $\{r Q^\star : r > 0\}$. The normalization in $G$ eliminates this scale confusion. Lemmas~\ref{support_lemma3_secondary}-\ref{support_lemma4_secondary} and the projective Lyapunov argument in Theorem~\ref{Risk:Thm3} demonstrate that $Q^\star$ is the unique globally asymptotically stable equilibrium of $\dot{Q}(t) = G(Q(t)) - Q(t)$. Third, Lemma~\ref{finite-positive} reformulates Eq.~\eqref{eq:Q} as
\begin{eqnarray*}
&Q_{n+1}(i, u, v) = Q_n(i, u, v) + a(\nu(i, u, v, n)) \nonumber \\
&\mathbb{1}_{\{X_n = i, U_n = u, V_n = v\}}\left(G_{iuv}(Q_n) - Q_n(i, u, v) + M_{n+1}(i, u, v)\right),
\end{eqnarray*}
where $\bE[M_{n+1}(i, u, v) | \pazocal{F}_n, X_n = i, U_n = u, V_n = v] = 0$. Positive initialization and $0 < a(n) < 1$ maintain strict positivity. On $D_{R, \delta} := \left\{Q > 0: \|Q\|_\infty \leqslant R, Q(i_0, u_0, v_0) \geqslant \delta \right\}$, Lemma~\ref{local_Lipschitzness} demonstrates that $G - I$ is Lipschitz and that the martingale-difference noise has uniformly bounded conditional second moment. Assumption~\ref{Viable_Safe} and the local-clock argument presented in Lemma~\ref{Stability} further give
\begin{equation*}
\sup_{n \geqslant 0} \|Q_n\|_\infty < \infty, \quad \liminf_{n \rightarrow \infty} Q_n(i_0, u_0, v_0) > 0.
\end{equation*}
The iterates then lie within a random compact set $D_{R, \delta}$. The stepsize and continuous sampling conditions enable the application of the asynchronous ODE theorem to the localized recursion. Combining the resulting ODE tracking with Theorem~\ref{Risk:Thm3} gives $Q_n \rightarrow Q^\star \quad \text{and} \quad Q_n(i_0, u_0, v_0) \rightarrow \Lambda^\star.$
\subsection{Convergence of Perron-Frobenius Eigen Estimates}\label{eigenpair_convergence}
We now explain the convergence of the eigenpair recursion in Eqs.~\eqref{principal_eigenvector}-~\eqref{principal_eigenvalue}. This step builds upon a block-wise implementation of the Perron-Frobenius stochastic-approximation method in Section~VI of \cite{borkar2014asynchronous}. For each fixed action pair $(u, v) \in \U \times \V$, recall that $\bP^0_{u,v}$ denotes the restriction of $\bP_{u,v}$ to $\X_0$. According to Assumption~\ref{assume:irr}, when applied to the deterministic stationary policy pair that chooses $(u, v)$ at every state, $\bP^0_{u,v}$ is an irreducible nonnegative matrix within the finite state space $\X_0$. Furthermore, $d_{u, v}(i) > 0$ ensures that no safe row is degenerate. Conditional on $\pazocal{F}_n$ and $\{X_n = i, U_n = u, V_n = v\}$, the sample term in Eq.~\eqref{principal_eigenvector} satisfies
\begin{align*}
&\bE[\mathbb{1}_{\{X_{n+1} \in \X_0\}} \e_n(X_{n+1}, u, v) | \pazocal{F}_n, X_n = i, U_n = u, V_n = v]\\
&= \sum_{j \in \X_0} \bP(j | i, u, v) \e_n(j, u, v) = (\bP^0_{u,v} \e_n(\cdot, u, v))(i).
\end{align*}
Thus, Eq.~\eqref{principal_eigenvector} is the asynchronous stochastic-approximation equivalent of normalized Perron-Frobenius iteration
\begin{equation*}
\dot{\e}(\cdot, u, v) = \frac{\bP^0_{u,v} \e(\cdot, u, v)}{\alpha^\top \e(\cdot, u, v)} - \e(\cdot, u, v),
\end{equation*}
up to the standard positive diagonal time-rescaling resulting from asynchronous component updates. Unsafe transitions only add a zero sample to the substochastic row, thereby not affecting the conditional mean field above. Following the normalization of the prescribed positive weight vector $\alpha$, the assumptions presented in Section~VI of \cite{borkar2014asynchronous} are applicable to each block $(u, v)$: the matrix $\bP^0_{u,v}$ is irreducible and nonnegative, the initial vector is componentwise positive, the stepsizes satisfy the Robbins-Monro conditions, and Assumption~\ref{Viable_Safe} ensures that every component $(i, u, v)$ is updated with asymptotically positive frequency. Therefore, utilizing the synchronous and asynchronous Perron-Frobenius convergence results from Theorems~7 and~8 in \cite{borkar2014asynchronous}, applied individually to each finite block $(u, v)$ and subsequently intersecting the corresponding probability one events over $\U \times \V$, we derive
\begin{equation*}
\e_n(\cdot, u, v) \rightarrow \e(\cdot, u, v), \Lambda_n(u, v) = \alpha^\top \e_n(\cdot, u, v) \rightarrow \Lambda(u, v), 
\end{equation*}
\text{a.s.} simultaneously for all $(u, v) \in \U \times \V$, where for action pair $(u, v)$, $\Lambda$ denotes the principal eigenvalue of $\bP^0_{u,v}$ and
\begin{equation*}
\bP^0_{u,v} \e(\cdot, u, v) = \Lambda(u, v) \e(\cdot, u, v), \quad \alpha^\top \e(\cdot, u, v) = \Lambda(u, v).
\end{equation*}
Thus Perron-Frobenius eigenpair estimates used for the empirical twisted kernel $\up_n$ converge almost surely.
\subsection{Convergence of Regular Control} \label{Convergence_Regular}
We finally justify the convergence of the slow regular-control recursion in Eq.~\eqref{eq:Q-regular}. %
Upon the convergence of the fast variables, Eq.~\eqref{eq:Q-regular} transforms into an asynchronous relative Q-learning recursion for a finite-state average-cost MDP, as studied by \cite{abounadi2001learning}. We first demonstrate that the emergency action observed by the regular recursion has a limiting conditional distribution. According to Theorem~\ref{emergency_convergence}, $Q_n \rightarrow Q^\star$ almost surely. Since $\U$ and $\V$ are finite, for every $i \in \X_0$ and $v \in \V$,
\begin{equation*}
\min_{u \in \U} Q_n(i, u, v) \rightarrow \min_{u \in \U} Q^\star(i, u, v) \quad \text{a.s.}
\end{equation*}
Fix a sample path on which this convergence holds. Given that
\begin{equation*}
|\min_{u \in \U} (Q_n(i, u, v) - Q^\star(i, u, v))| \leqslant \max_{u \in \U}|Q_n(i, u, v) - Q^\star(i, u, v)|
\end{equation*}
it follows that as $n \rightarrow \infty$, the right-hand side approaches $0$, indicating that the left-hand side also approaches $0$. Thus, for any positive value $\frac{\Delta}{4}$, there exists a finite time $N$ such that  
\begin{eqnarray*}
\big|\min_{u \in \U} Q_n(i, u, v) - \min_{u \in \U} Q^\star(i, u, v)\big| < \frac{\Delta}{4}, \text{~for all~} n \geqslant N.
\end{eqnarray*}
If $\bar{v}_n(i)$ changes after this time, then according to the update rule in Algorithm~\ref{algo:Q}, the newly chosen action, $v^+$, satisfies
\begin{eqnarray*}
&\min_{u \in \U} Q_n(i, u, v^+) > \min_{u \in \U} Q_n(i, u, \bar{v}_n(i)) + \Delta.\\
\text{~Thus} &\min_{u \in \U} Q^\star(i, u, v^+) > \min_{u \in \U} Q^\star(i, u, \bar{v}_n(i)) + \frac{\Delta}{2}.
\end{eqnarray*}
Thus, after a finite time, every change of $\bar{v}_n(i)$ strictly improves the limiting safety score by at least $\frac{\Delta}{2}$. Given that $\V$ is finite, only a finite number of such changes can happen. Thus, $\bar{v}_n(i)$ converges almost surely for every $i \in \X_0$. Therefore, given that $X_n = i$, the distribution of $V_n$ converges to
\begin{equation*}
(1 - \epsilon) \mathbb{1}_{\{v = \lim_{\ell \rightarrow \infty} \bar{v}_\ell(i)\}} + \frac{\epsilon}{|\V|}, \quad v \in \V.
\end{equation*}
The remaining randomness in $V_n$ is thus ordinary sampling noise around this limiting distribution, rather than a time-varying drift. Using the Perron-Frobenius limits in Section~\ref{eigenpair_convergence}, the limiting equivalent of the empirical twisted kernel in Eq.~\eqref{eq:twist-2}, within the conditional mean of the slow recursion, is
\begin{eqnarray*}
&\sum_{v \in \V} \Big((1 - \epsilon) \mathbb{1}_{\{v = \lim_{\ell \rightarrow \infty} \bar{v}_\ell(i)\}} + \frac{\epsilon}{|\V|}\Big) \mathbb{1}_{\{j \in \X_0\}} \bp(j | i, u, v)\\
&\cdot \frac{\e(j, u, v)}{\Lambda(u, v) \e(i, u, v)}, \quad i, j \in \X_0, u \in \U.
\end{eqnarray*}
For each fixed pair $(u, v)$, the Perron-Frobenius relation
\begin{eqnarray*}
&\sum_{j \in \X_0} \bp(j | i, u, v) \e(j, u, v) = \Lambda(u, v) \e(i, u, v)\text{~implies that}\\
&\sum_{j \in \X_0} \mathbb{1}_{\{j \in \X_0\}} \bp(j | i, u, v) \frac{\e(j, u, v)}{\Lambda(u, v) \e(i, u, v)} = 1.
\end{eqnarray*}
Consequently, the limiting equivalent of $\breve{p}_n(\cdot | i, u)$, after averaging over the limiting conditional distribution of $V_n$, is a probability distribution on $\X_0$. Furthermore, as each $\e(\cdot, u, v)$ is componentwise strictly positive, the twisting operation maintains the support of the corresponding safe kernel. Thus, under any stationary regular policy $\phi$, the Markov chain derived from this limiting twisted kernel has the same support as the safe kernel derived from $\phi$ together with the limiting emergency action distribution above. The irreducibility and unichain condition required for average-cost Q-learning are consequently derived from Assumption~\ref{assume:irr}. The multiplicative factor in Eq.~\eqref{eq:Q-regular} is exactly the likelihood-ratio modification related to this twisted kernel. Conditional on $\pazocal{F}_n$ and on $\{X_n = i, U_n = u, V_n = v\}$, the limiting conditional mean of the transition term in Eq.~\eqref{eq:Q-regular} satisfies $\bE\big[\mathbb{1}_{\{X_{n+1} \in \X_0\}}\cdot $
\begin{align*}
&\frac{\e(X_{n+1}, u, v)}{\Lambda(u, v) \e(i, u, v)} \min_{u' \in \U} \widetilde{Q}_n(X_{n+1}, u') | \pazocal{F}_n, X_n, U_n, V_n\big]
\\
&= \sum_{j \in \X_0} \mathbb{1}_{\{j \in \X_0\}} \bp(j | i, u, v) \frac{\e(j, u, v)}{\Lambda(u, v) \e(i, u, v)} \min_{u' \in \U} \widetilde{Q}_n(j, u').
\end{align*}
Averaging over limiting conditional distribution of $V_n$ yields
\begin{align*}
&\bE\Big[\mathbb{1}_{\{X_{n+1} \in \X_0\}} \frac{\e(X_{n+1}, u, V_n)}{\Lambda(u, V_n) \e(i, u, V_n)} \min_{u' \in \U} \widetilde{Q}_n(X_{n+1}, u') |\\
&\pazocal{F}_n, X_n = i, U_n = u\Big]\longrightarrow \sum_{j \in \X_0} \sum_{v \in \V} \Big((1 - \epsilon)\mathbb{1}_{\{v = \lim_{\ell \rightarrow \infty} \bar{v}_\ell(i)\}} 
\\
& + \frac{\epsilon}{|\V|}\Big) \mathbb{1}_{\{j \in \X_0\}} \bp(j | i, u, v) \frac{\e(j, u, v)}{\Lambda(u, v) e(i, u, v)} \min_{u' \in \U} \widetilde{Q}_n(j, u').
\end{align*}
Thus, the additional factor in Eq.~\eqref{eq:Q-regular} does not change the nature of the average-cost Q-learning recursion. It changes the transition kernel from the original safe kernel to the limiting twisted kernel as indicated by Eq.~\eqref{eq:twist-2}. Consequently, the limiting mean field of Eq.~\eqref{eq:Q-regular} is
\begin{align*}
\dot{\widetilde{Q}}(i, u) &= \sum_{j \in \X_0} \sum_{v \in \V} \Big((1 - \epsilon) \mathbb{1}_{\{v = \lim_{\ell \rightarrow \infty} \bar{v}_\ell(i)\}} + \frac{\epsilon}{|\V|}\Big)\\
&\cdot \mathbb{1}_{\{j \in \X_0\}} \bp(j | i, u, v) \frac{\e(j, u, v)}{\Lambda(u, v) \e(i, u, v)} \min_{u' \in \U} \widetilde{Q}(j, u') \\
&+ w(i, u) - \min_{u' \in \U} \widetilde{Q}(i_0, u') - \widetilde{Q}(i, u).
\end{align*}
This is the limiting ODE of asynchronous RVI Q-learning for the finite-state average-cost MDP with state space $\X_0$, action space $\U$, one-stage cost $w(i, u)$, transition probabilities given by the limiting equivalent of Eq.~\eqref{eq:twist-2}, and relative normalization $\min_{u' \in \U} \widetilde{Q}(i_0, u')$. This normalization is one of the admissible normalizations in \cite{abounadi2001learning}. The convergence theorem for RVI Q-learning in \cite{abounadi2001learning} applies to both synchronous and asynchronous implementations once boundedness is available.
{\cred The following convergence argument assumes that $\{\widetilde{Q}_n\}$ is almost surely bounded.} It remains to account for the fact that Eq.~\eqref{eq:Q-regular} uses the online estimates $(\e_{n+1}, \Lambda_{n+1})$, rather than their limits. This produces a perturbation that asymptotically approaches zero. From Section~\ref{eigenpair_convergence},
\begin{equation*}
\e_n(\cdot, u, v) \rightarrow \e(\cdot, u, v), \quad \Lambda_n(u, v) \rightarrow \Lambda(u, v) \quad \text{a.s.}
\end{equation*}
for every $(u, v)$. Given that $\X_0$, $\U$, and $\V$ are finite, and that $\e(\cdot, u, v)$ and $\Lambda(u, v)$ are strictly positive, the denominators in Eq.~\eqref{eq:Q-regular} are almost surely eventually bounded away from zero. Thus, the difference between the online conditional mean and the limiting conditional mean above converges uniformly to zero for all $(i, u)$. 
{\cred The difference between current conditional mean and the actual update results in martingale-difference noise.}
Consequently, according to the convergence theorem for synchronous and asynchronous RVI Q-learning in Theorem~3.5 of \cite{abounadi2001learning}, 
the regular-control iterates converge almost surely:
$\widetilde{Q}_n \rightarrow \widetilde{Q}^\star \quad \text{a.s.}$
The limit $\widetilde{Q}^\star$ solves the average-cost optimality equation
\begin{align*}
&\widetilde{Q}^\star(i, u) = \sum_{j \in \X_0} \sum_{v \in \V} \Big((1 - \epsilon) \mathbb{1}_{\{v = \lim_{\ell \rightarrow \infty} \bar{v}_\ell(i)\}} + \frac{\epsilon}{|\V|}\Big)\mathbb{1}_{\{j \in \X_0\}}\\
&\bp(j | i, u, v) \frac{\e(j, u, v)}{\Lambda(u, v) \e(i, u, v)} \min_{u' \in \U} \widetilde{Q}^\star(j, u')+ w(i,u) - \beta.
\end{align*}
under the specified normalization. Thus, any greedy selector
\begin{equation*}
\phi^\star(i) \in \arg\min_{u \in \U} \widetilde{Q}^\star(i, u)
\end{equation*}
is optimal for the regular-control problem associated with the limiting twisted kernel obtained by Eq.~\eqref{eq:twist-2}.
\section{Simulations}
\label{sec:simulations}
\begin{figure}[t]
    \centering
    \includegraphics[width=\linewidth]{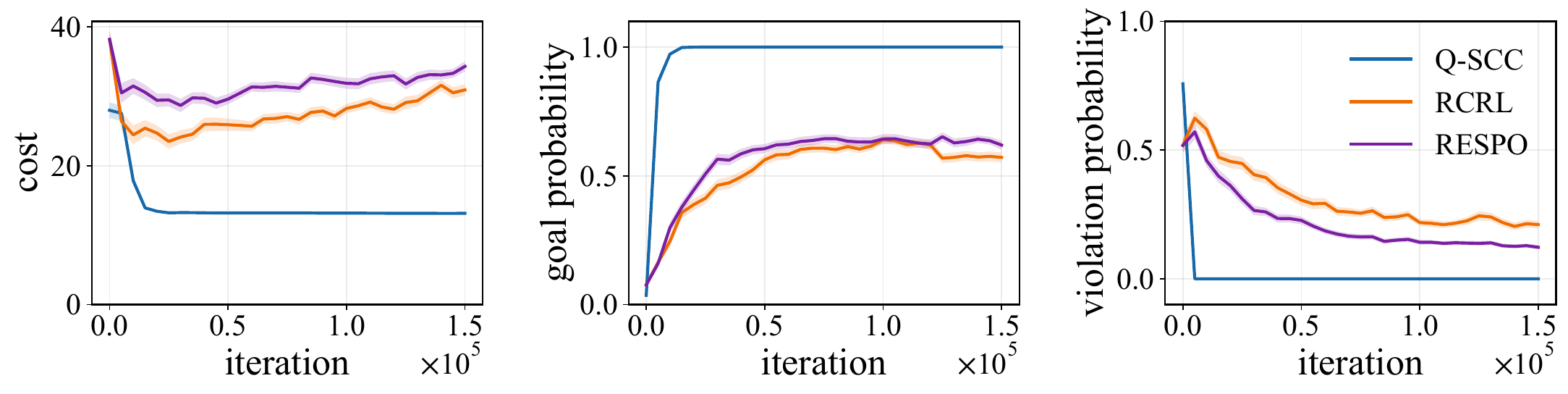}
    \caption{Training plots for GridWorld. From left to right, the plots show cost, goal probability, and violation probability.}
    \label{fig:gridworld-learning}
\end{figure}

\begin{figure}[t]
    \centering
    \includegraphics[width=\linewidth]{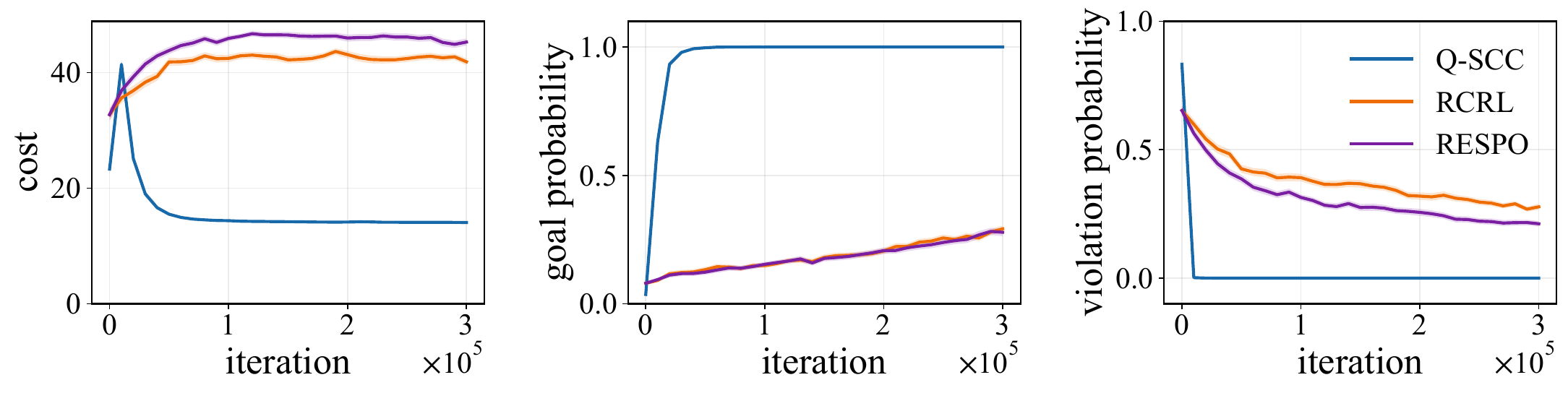}
    \caption{Training plots for SafetyCarGoal. From left to right, the plots show cost, goal probability, and violation probability.}
    \label{fig:cargoal-learning}
\end{figure}

This section presents the experimental evaluation of our proposed approaches in two stochastic finite-state environments: a GridWorld and a fixed-layout CarGoal task derived from Safety Gymnasium. We considered two benchmark approaches that considered state safety: (i)~RCRL is a Lagrangian-based algorithm with a safety value function to maintain the safety constraint within its estimated feasible set~\cite{yu2022reachability}, and (ii)~RESPO algorithm estimates the probability of a future constraint violation and balances the maximization of rewards with the minimization of cumulative safety costs~\cite{ganai2024iterative}.
{\cred All methods share the regular and emergency actions, safety information, transition sampler, and task cost. Our approach optimizes average task cost under the transformed safe dynamics, while the baselines use discounted objectives.}
We evaluated learned policies in terms of safety and meeting the goal. Each method performs training for $150000$ iterations in GridWorld and $300000$ iterations in CarGoal, utilizing the same $100$ random seeds across all methods within each environment. Our proposed approach utilizes a fixed exploration probability of $0.10$. 
{\cred Each iteration is one training transition. The results represent averages derived from the independent training iterations associated with the same $100$ seeds. Error bars and shaded regions represent one standard deviation of the mean. Policies are evaluated from $i_0$ by propagating probabilities through the finite model until the first goal or unsafe state, or $60$ decisions. Goal and violation probabilities refer to the corresponding terminal occurrences within this horizon. Cost is the expected cumulative $w$ before termination. Exploration has been deactivated, but transition randomness remains. Our approach uses $\arg\min_u \widetilde{Q}_n(i, u)$ and the stored $\bar{v}_n(i)$, while the baselines select their most probable joint action. In CarGoal, one macro-action is one decision. Figures~\ref{fig:gridworld-learning} and~\ref{fig:cargoal-learning} evaluate checkpoint policies, rather than counting violations during exploratory training trajectories.}

\begin{figure}
\centering
\subcaptionbox{\label{fig:gridworld-layout}}{\includegraphics[scale=0.16]{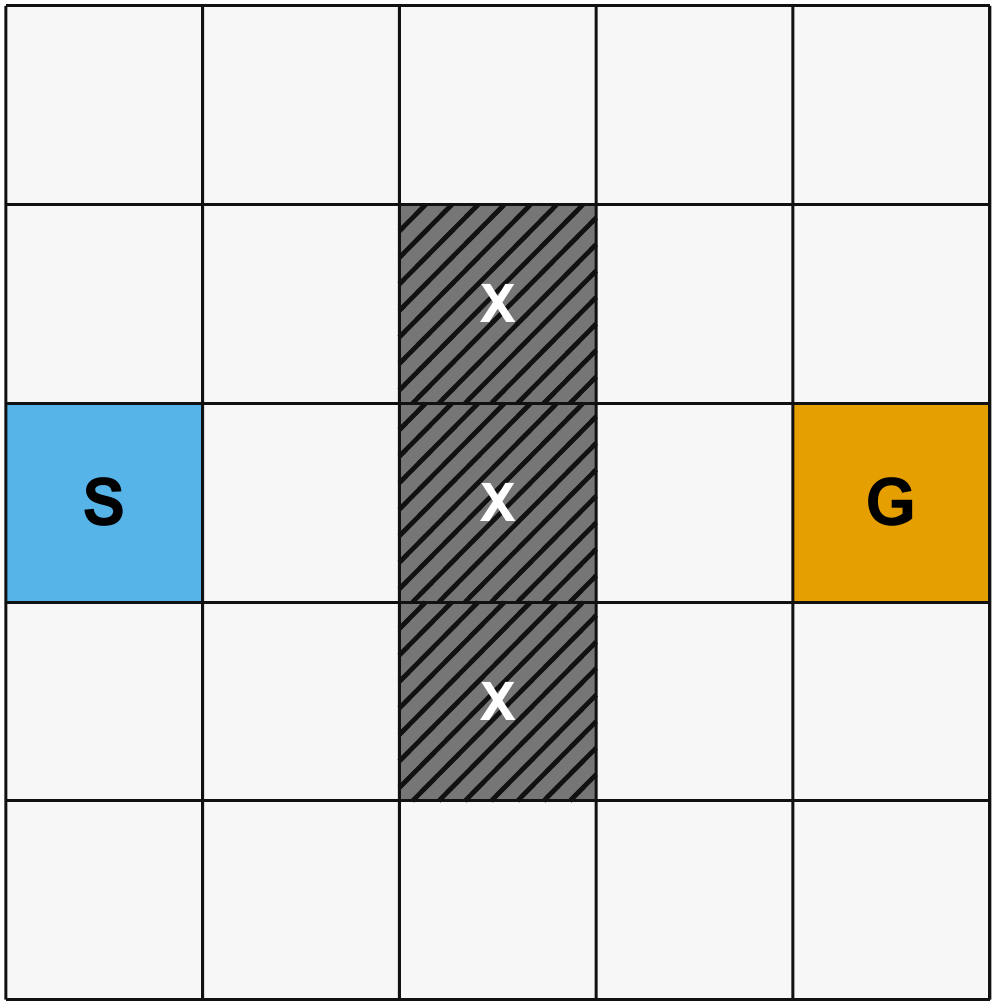}}
\subcaptionbox{\label{fig:cargoal-layout_1}}{\includegraphics[scale=0.12]{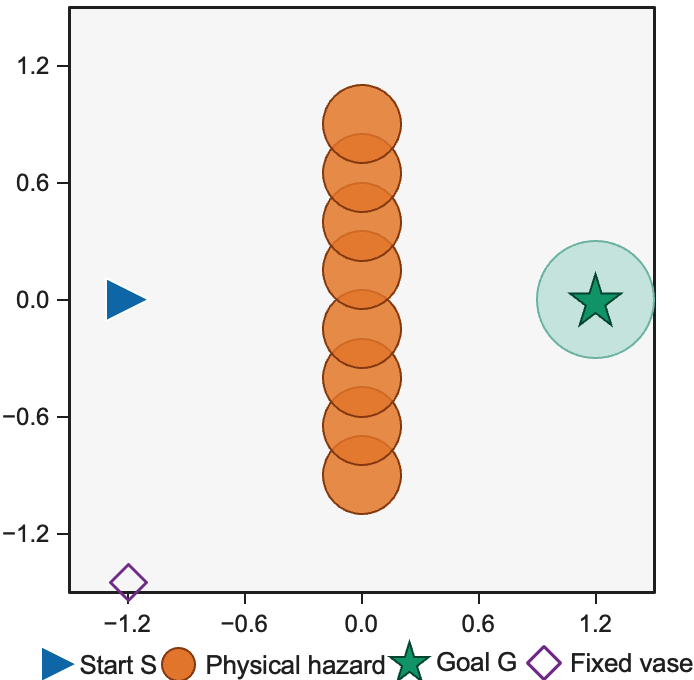}}
\hspace{-0.3 em}%
\subcaptionbox{\label{fig:cargoal-layout_2}}{\includegraphics[scale=0.12]{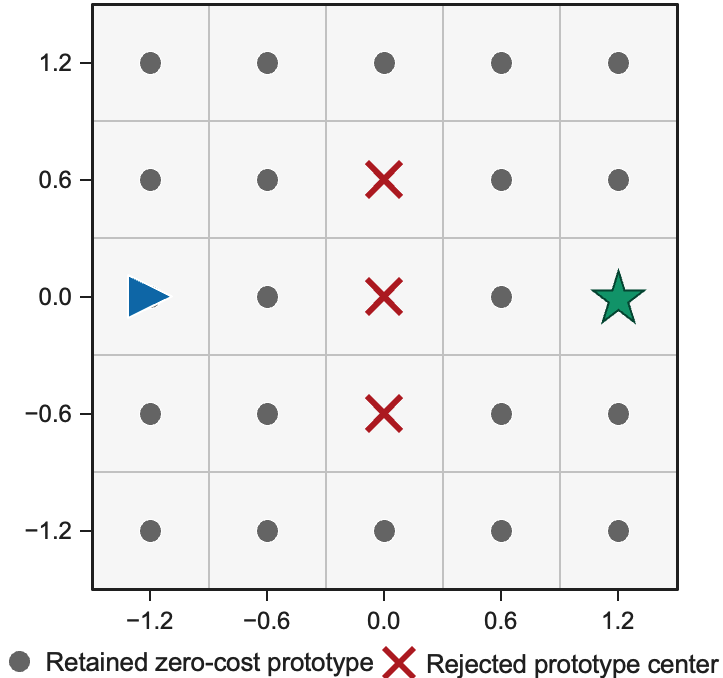}}
\caption{\footnotesize Figure~\ref{fig:gridworld-layout} is the Stochastic GridWorld layout. The agent starts at $S$, aims to reach $G$, and must avoid the three unsafe cells. Figures~\ref{fig:cargoal-layout_1} and~\ref{fig:cargoal-layout_2} shows the Fixed-layout SafetyCarGoal environment and its canonical finite model layout. Figure~\ref{fig:cargoal-layout_1}: The physical layout includes the start point $S$, the goal point $G$, eight overlapping native hazard disks, and one fixed vase. Figure~\ref{fig:cargoal-layout_2}: The positional grid includes $25$ canonical centers. Three crosses denote the rejected center points, while the other $21$ non-goal central points are maintained. Every maintained center point is expanded into four headings and two speeds, while the goal point is represented by one aggregate state.} \label{fig:gridworld-final}
\end{figure}
\textbf{Stochastic GridWorld.} We analyze the $5 \times 5$ GridWorld presented in Figure~\ref{fig:gridworld-final}. The agent starts at $i_{0} = (2, 0)$ and is tasked with reaching $i_{\mathrm{goal}} = (2, 4)$. The cells $(1, 2)$, $(2, 2)$, and $(3, 2)$ are considered hazardous. The remaining $22$ cells constitute the safe set $\X_0$, with all unsafe cells represented by a common unsafe state $i_{\mathrm{unsafe}}$. As a result, the tabular model includes $23$ states.
In every safe state, the regular controller chooses a direction $u \in \{\mathrm{N}, \mathrm{E}, \mathrm{S}, \mathrm{W}\}$ and the emergency controller selects a mode $v \in \{\mathrm{Shield},\mathrm{Normal}\}$.
{\cred At non-goal states, for an action chosen from $\{N, E, S, W\}$, both environments assign probabilities $(0.6, 0.12, 0.12, 0.06, 0.1)$ under Shield and $(0.7, 0.1, 0.1, 0.05, 0.05)$ under Normal to movement in the selected direction, $90^\circ$ to its left, $90^\circ$ to its right, in the opposite direction, and staying in place, respectively. Both finite models use these probabilities and satisfy $d_{u, v}(i) \geqslant 0.3$.}
An action that exceeds the grid boundary results in the agent remaining in the current cell. When a sampled regular control action suggests a transition to an unsafe state $j$, $\mathrm{Shield}$ maintains the agent in its current safe cell, while $\mathrm{Normal}$ moves the agent to $j$. Thus, $\bP(j\in \X\setminus\X_0 | i, u, \mathrm{Shield}) = 0$ for $i \in \X_0$, while the $\mathrm{Normal}$ mode may violate safety. The per-stage cost is $w(i, u) = 1$ for all non-goal safe states and $w(i_{\mathrm{goal}}, u) = 0$ for the goal state. 
{\cred Shield does not impose a separate penalty. However, its lower probability of achieving the desired movement may delay goal reaching and increase task cost.}
The training kernel restarts from a safe state upon reaching either $i_{\mathrm{goal}}$ or $i\in \X\setminus \X_0$.

\textbf{Modified Safety Gym CarGoal.} We develop a fixed-layout, canonical finite macro-MDP derived from the original SafetyCarGoal1-v0 simulator in Safety Gymnasium.
The chosen physical domain is $\mathcal{D} = [- 1.5, 1.5]^2$, the agent starts at $(- 1.2, 0)$, and its goal is located at $(1.2, 0)$. Eight native hazard disks with a radius of $0.2$ are positioned along the line $x = 0$, forming a central barrier flanked by upper and lower corridors, as shown in Figures~\ref{fig:cargoal-layout_1} and ~\ref{fig:cargoal-layout_2}. A native vase is positioned at the coordinates $(-1.2, -1.45)$. We establish a $5 \times 5$ canonical position grid at the coordinates $\{- 1.2, - 0.6, 0, 0.6, 1.2\}^2$. At every positional center, the native simulator performs an evaluation at four directions $\{0, \frac{\pi}{2}, \pi, - \frac{\pi}{2}\}$ and two speeds $\{0, 0.08\}$. The three points $(0, -0.6)$, $(0, 0)$, and $(0, 0.6)$ are excluded. The remaining $21$ non-goal centers are each developed into eight position-heading-speed prototypes, while all goal-achieving configurations are encapsulated by a singular aggregate goal state. As a result, the model includes $21 \times 4 \times 2 + 1 = 169$ safe states and one unsafe state.
An executed macro-transition is mapped to the aggregate unsafe state if its native trajectory results in a positive safety cost or exits $\mathcal{D}$. Every canonical prototype maintains the associated $72$-dimensional native observation. Thus, this represents a simulator-driven finite abstraction. In every non-goal safe state, the regular controller provides a direction $u \in \{\mathrm{N}, \mathrm{E}, \mathrm{W}, \mathrm{S}\}$, while the emergency controller chooses $v \in \{\mathrm{Shield}, \mathrm{Normal}\}$.
{\cred Each macro-action starts from its prescribed position, heading, and speed. Shield blocks hazardous targets by staying in place and follows safe waypoints. Normal tries the target directly. }
The goal can be reached within a finite number of safe macro choices. The cost of a per-stage task is $w(i, u) = 1$ for all non-goal safe states and $w(i_{\mathrm{goal}}, u) = 0$, indicating that a lower cost in safe policies correlates with faster goal achievement in macro decisions.
\begin{figure}
\centering
\subcaptionbox{\label{fig:1}}{\includegraphics[scale=0.23]{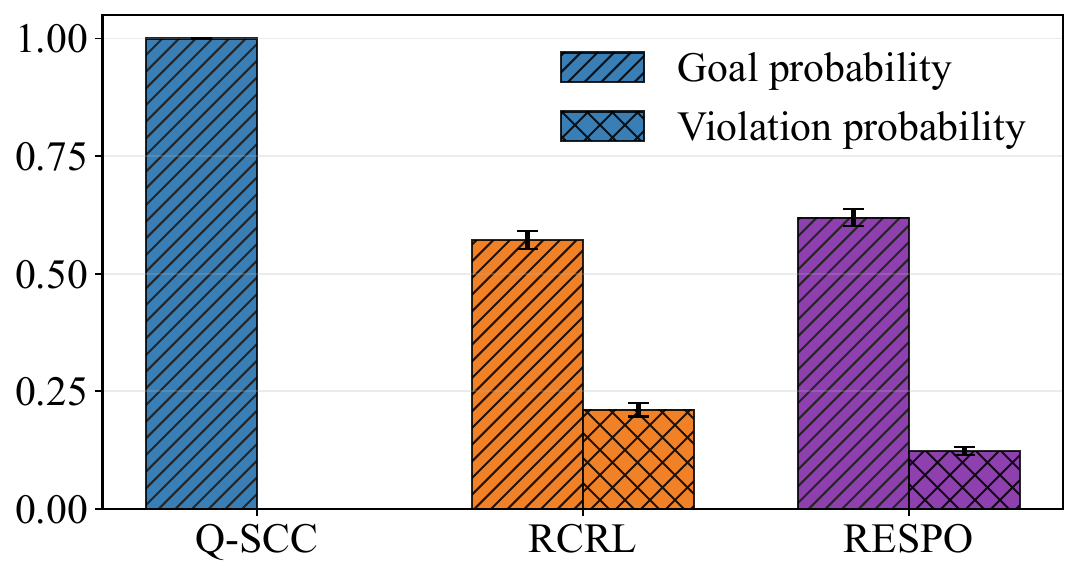}}
\hspace{0.3 em}%
\subcaptionbox{\label{fig:2}}{\includegraphics[scale=0.23]{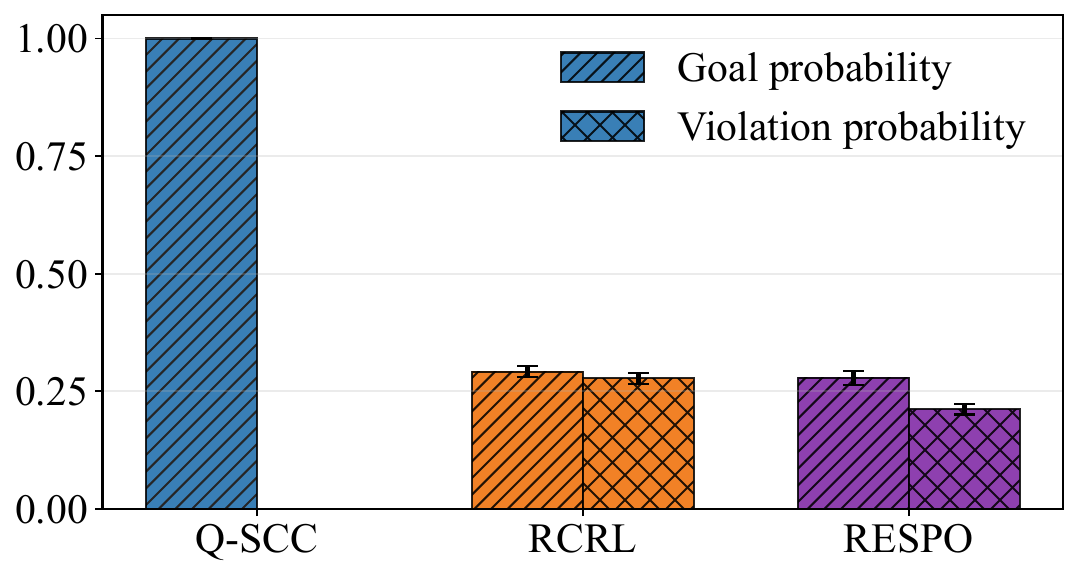}}
\caption{\footnotesize Final policy evaluation. Figure~\ref{fig:1} illustrates the probabilities of reaching the goal and the probability of entering an unsafe state for gridworld, while Figure~\ref{fig:2} refers to cargoal.} \label{fig:gridworld-final_2}
\end{figure}

Figures~\ref{fig:1} and~\ref{fig:2} compare the evaluation of the learned policies. Our proposed algorithm achieves both a substantial goal probability and a precisely zero violation probability. In contrast, both the RCRL and RESPO methods yield a lower goal probability and exhibit high violation probability. 
Figures~\ref{fig:gridworld-learning} and~\ref{fig:cargoal-learning} demonstrate the comparison during learning. 
{\cred In GridWorld, the eight-step shortest safe path assumes intended moves, while stochastic deviations and stays increase the expected number of decisions.}
The plots show that our proposed algorithm consistently outperforms other benchmark algorithms.

{\cred The safe kernels are irreducible in GridWorld but reducible in CarGoal. The latter experiment thus falls beyond Assumption~\ref{assume:irr}. Shield provides $\Lambda^\star = 1$ in both models. For a fixed Shield action, the safe kernel is stochastic and has an eigenvalue of one with a consistent eigenvector. With this eigenpair and $V_n = \mathrm{Shield}$, Eq.~\eqref{eq:Q-regular} reduces to RVI. Our approach also learns which emergency action to select.}
\appendix
\subsection{Proving $Q^\star$ is a fixed point of ODE}
\begin{lemma}\label{logsumexp}
Let $x_{\max}=\max\{x_1,\cdots,x_n\}$ and define $\logsumexp(x)=\log\sum_{i=1}^{n}e^{x_i}$. Then, for any $a>0$, 
$
\lim_{a\rightarrow\infty}\frac{1}{a}\logsumexp(ax)=x_{\max}.
$
\end{lemma}
\begin{proof}
We begin by noting that
$e^{x_{\max}} \leqslant \sum_{i=1}^{n}e^{x_i} \leqslant \sum_{i=1}^{n}e^{x_{\max}}=ne^{x_{\max}}.$
Taking the logarithm of all sides of this inequality, we obtain $x_{\max} \leqslant \logsumexp(x) \leqslant x_{\max} + \log{n}.$
Now, consider any $a>0$. By applying this inequality to $ax$, $ax_{\max} \leqslant \logsumexp(ax) \leqslant ax_{\max} + \log{n}.$ 
Dividing by $a$ yields $x_{\max} \leqslant \frac{1}{a}\logsumexp(ax) \leqslant x_{\max} + \frac{1}{a}\log{n}.$
Finally, by taking the limit as $a\rightarrow\infty$, we conclude that $\lim_{a\rightarrow\infty}\frac{1}{a}\logsumexp(ax)=x_{\max}.$
\end{proof}
Moreover, by Lemma~\ref{logsumexp} and Definition~\ref{def:Th}, we have $(\widehat{T}x)(i)$
\begin{align*}
&=\lim_{a\rightarrow\infty}\frac{1}{a}\max_{v}\min_{u}(c_{uv}(i)+\log\sum_{j\in\X_0}e^{ax(j)+\log{P_{uv}^{\star}(j|i)}})\\
&=\max_{v}\min_{u}\max_{j\in\mathrm{supp}(P_{uv}^{\star}(\cdot|i))}x(j),
\end{align*}
where $\mathrm{supp}(P_{uv}^{\star}(\cdot|i)) := \{j \in \X_0: P_{uv}^{\star}(j|i) > 0\}$ is the support. Thus, the limit in the definition of $\widehat{T}$ exists for every $x$. 
Finally, since $Q^\star$ is a fixed point of $F$, we have $Q^{\star} = F(Q^\star)$, and hence, for any $(i,u,v)$, $\frac{F_{iuv}(q)}{Q^{\star}(i,u,v)} = \frac{F_{iuv}(q)}{F_{iuv}(Q^\star)}$. 

Since the positive fixed point of $F$ is unique only up to multiplication by a positive scalar, we fix its representative by enforcing the normalization $Q^\star(i_0, u_0, v_0) = \Lambda^\star$. Then, $Q^\star$ represents an equilibrium of the normalized operator $G$. Indeed, $G(Q^\star) = \frac{\Lambda^\star}{Q^\star(i_0, u_0, v_0)} F(Q^\star) = F(Q^\star) = Q^\star$.

\begin{lemma}\label{fixed point}
Let $x\in\bR^{|\X_0|}$ be such that $\widehat{T}x=x$. Then $x$ is constant, i.e., $x \equiv c\mathbf{1}$ for some $c \in \bR$. In particular, $\widehat{T}$ has only trivial fixed points.
\end{lemma}
\begin{proof}
Recall that for each $i \in \X_0$, $(\widehat{T}x)(i)=\max_{v}\min_{u}\max_{j\in\mathrm{supp}(P_{uv}^{\star}(\cdot|i))}x(j).$
First observe that every constant vector is a fixed point of $\widehat{T}$. If $x \equiv c \mathbf{1}$ for some $c \in \bR$, then for all $(u,v)$ and all $i \in \X_0$, $\max_{j\in\mathrm{supp}(P_{uv}^{\star}(\cdot|i))}x(j)=c,$
and hence $(\widehat{T}x)(i)=c=x(i)$.
We now show that these are the only fixed points. For contradiction, suppose that $x$ is a fixed point of $\widehat{T}$ and that entries of $x$ are not constant. Define $m:=\min_{k}x(k), \quad S:=\{k:x(k)=m\}.$
Then $S$ is nonempty and, since $x$ is not constant, $S$ is a proper subset of $\X_0$. 
Fix any $i \in S$, so that $x(i)=m$. For each $(u,v)$, $g(i,u,v)=\max_{j\in\mathrm{supp}(P_{uv}^{\star}(\cdot|i))}x(j).$
Since $x(j) \geqslant m$ for all $j$, we have $g(i,u,v) \geqslant m$ for all $(u,v)$. Since $x$ is a fixed point of $\widehat{T}$, we obtain $m=x(i)=(\widehat{T}x)(i)=\max_{v}\min_{u}g(i,u,v).$
Thus there exists $v_{1}(i)$ achieving the maximum and, for this $v_{1}(i)$, some $u_{1}(i)$ achieving the minimum, such that $g(i,u_{1}(i),v_{1}(i))=m.$
By the definition of $g$, we have $m=g(i,u_{1}(i),v_{1}(i))=\max_{j\in\mathrm{supp}(P_{u_{1}(i)v_{1}(i)}^{\star}(\cdot|i))}x(j).$
Since all $x(j) \geqslant m$, the fact that this maximum equals $m$ implies that $x(j)=m \quad \text{for all} \quad j\in\mathrm{supp}(P_{u_{1}(i)v_{1}(i)}^{\star}(\cdot|i)).$
Equivalently, $\mathrm{supp}(P_{u_{1}(i)v_{1}(i)}^{\star}(\cdot|i)) \subseteq S,$
and hence
\begin{equation*}
P_{u_{1}(i)v_{1}(i)}^{\star}(S|i)=1, \quad P_{u_{1}(i)v_{1}(i)}^{\star}(\X_0 \setminus S|i)=0.
\end{equation*}
Using $P_{uv}^{0}(j|i)=d_{uv}(i)P_{uv}^{\star}(j|i)$, we further derive
\begin{equation*}
P_{u_{1}(i)v_{1}(i)}^{0}(S|i)=d_{u_{1}(i)v_{1}(i)}, \quad P_{u_{1}(i)v_{1}(i)}^{0}(\X_0 \setminus S|i)=0.
\end{equation*}
Since $i \in S$ was arbitrary, we can define a pair of controls $(u_{1}(\cdot),v_{1}(\cdot))$ on $S$ such that $P_{u_{1}(i)v_{1}(i)}^{0}(\X_0 \setminus S|i)=0$ for every $i \in S$. Thus $S$ is an absorbing subset for the kernel $\bP^0$ under this pair of controls, which contradicts Assumption~\ref{assume:irr}. Therefore $x$ must be a constant vector. We conclude that every fixed point of $\widehat{T}$ is a constant vector.
\end{proof}

\begin{theorem}\label{Risk:Thm1}
There exists a scalar $\lambda \in \bR$ and a vector $h \in \bR^{|\X_0|}$ such that $Th = \lambda \mathbf{1} + h$. 
\end{theorem}
\begin{proof}
By Lemma~\ref{fixed point}, the recession function $\widehat{T}$ of $T$ exists and has only trivial eigenvectors. Hence, we apply Theorem~13 in \cite{gaubert2004perron}, which demonstrates that if the recession function of a topical function has only trivial eigenvectors, then all slice spaces $S_{\alpha}^{\beta}(T):=\{x \in \bR^{|\X_0|}: \alpha \mathbf{1} + x \leqslant Tx \leqslant \beta \mathbf{1} + x\}$ are bounded in the Hilbert semi-norm.
Fix $\alpha = \min_{i \in \X_0}(T0)(i)$ and $\beta = \max_{i \in \X_0}(T0)(i)$. For any $c \in \bR$, consider the constant vector $x=c\mathbf{1}$. Using the translation invariance of $T$, we derive $T(x)=T(c\mathbf{1})=T(0+c\mathbf{1})=T(0)+c\mathbf{1}.$
By the definition of $\alpha$ and $\beta$, $\alpha \mathbf{1} + x \leqslant T(x) \leqslant \beta \mathbf{1} + x.$
Thus, $x \in S_{\alpha}^{\beta}(T)$, and in particular the entire line $\{c \mathbf{1}: c \in \bR\}$ is contained in $S_{\alpha}^{\beta}(T)$, so this slice space is nonempty. 
Slice space are invariant under $T$. Indeed, if $x \in S_{\alpha}^{\beta}(T)$, then $\alpha \mathbf{1} + x \leqslant Tx \leqslant \beta \mathbf{1} + x.$
Applying $T$ and using the monotonicity of $T$ together with translation invariance yields $T(\alpha \mathbf{1} + x) \leqslant T(Tx) \leqslant T(\beta \mathbf{1} + x).$
 That is, $\alpha \mathbf{1} + Tx \leqslant T^{2}x \leqslant \beta \mathbf{1} + Tx$. Hence $Tx \in S_{\alpha}^{\beta}(T)$. By induction, for any $x_0 \in S_{\alpha}^{\beta}(T)$, the entire orbit $\{T^{k}(x_0)\}$ is contained in the same slice space $S_{\alpha}^{\beta}(T)$. 
Since $S_{\alpha}^{\beta}(T)$ is bounded in the Hilbert semi-norm, every such orbit is bounded in the Hilbert semi-norm, that is, $\sup_{k \geqslant 0}\|T^{k}(x_0)\|_\H < \infty$. In particular, taking $x_0 = c \mathbf{1}$ for some $c \in \bR$, we obtain an orbit of $T$ that is bounded in the Hilbert semi-norm. 
By applying Theorem~9 in \cite{gaubert2004perron}, $T$ has an eigenpair $(\lambda,h)$ satisfying $Th=\lambda\mathbf{1}+h$, which completes the proof.
\end{proof}

\begin{lemma}\label{lambda_equal}
For $n \geqslant 1$ and $i \in \X_0$, define $J_{n}(i)=$
$$
\max_{v_{0:n-1}}\min_{u_{0:n-1}}\log\bE_{i}^{u_{0:n-1}v_{0:n-1}}\big[\exp\big(\sum_{t=0}^{n-1}c(X_t,U_t,V_t)\big)\big],$$
where the maximum and minimum are taken over admissible Markov decision rules on the horizon $0, \cdots, n-1$. Set $J_{0}\equiv0$. Fix any $i \in \X_0$ and define $\lambda^{\star}=\lim_{n\rightarrow\infty}\frac{1}{n}J_n(i)$. If $(\lambda,h)$ is an eigenpair of $T$, i.e., $Th = \lambda \mathbf{1} + h$, then $\lambda = \lambda^\star$.
\end{lemma}
\begin{proof}
Consider a horizon of length $n+1$ and a fixed pair of control sequences $(u_{0:n},v_{0:n})$. For any $i \in \X_0$, 
\begin{align*}
&\log\bE_i^{u_{0:n}v_{0:n}}\big[\exp\big(\sum_{t=0}^{n}c(X_t,U_t,V_t)\big)\big]\nn\\
&=\log\bE_i^{u_0=u,v_0=v}\big[\exp\big(c(i,u,v)\big)\exp\big(\sum_{t=1}^{n}c(X_t,U_t,V_t)\big)\big]\\
&=c(i,u,v) + \log\sum_{j\in\X_0}P_{uv}^{\star}(j|i)\bE_{j}^{u_{1:n}v_{1:n}}\big[\exp\big(\sum_{t=1}^{n}c(X_t,U_t,V_t)\big)\big]
\end{align*}
where in the last step we used the law of total expectation. 
Optimizing over the control sequences from time $t=1,\cdots,n$ starting from state $j$ yields $J_n(j)$, so after optimizing over the first-stage controls $u,v$ as well, it gives
\begin{eqnarray*}
&J_{n+1}(i)=\max_{v}\min_{u}\big\{c(i,u,v) + \log\sum_{j\in\X_0}P_{uv}^{\star}(j|i)\\
&\exp\big(J_n(j)\big)\big\}=(TJ_n)(i).
\end{eqnarray*}
Thus, by induction, $J_n=T^n J_0$ for any $n \geqslant 0$. 
Now, let $(\lambda,h)$ be an eigenpair of $T$, so that $Th=\lambda\mathbf{1}+h$. Using the translation invariance of $T$, we derive by induction $T^n h = h + n \lambda \mathbf{1}$ for any $n \geqslant 0$. 
We compare $J_n$ with this eigenvector $h$. Since $J_0 \equiv 0$, define $\alpha := \min_{k \in \X_0} (J_0(k) - h(k)) = - \max_{k \in \X_0} h(k)$ and $\beta := \max_{k \in \X_0} (J_0(k) - h(k)) = - \min_{k \in \X_0} h(k)$. Then componentwise $\alpha \mathbf{1} + h \leqslant J_0 \leqslant \beta \mathbf{1} + h.$ 
Using the monotonicity and translation invariance of $T$, we prove by induction that for all $n \geqslant 0$, 
\begin{equation}\label{Jn_bound}
\alpha \mathbf{1} + h + n \lambda \mathbf{1} \leqslant J_n \leqslant \beta \mathbf{1} + h + n \lambda \mathbf{1}.
\end{equation}
The case $n = 0$ holds by construction. Suppose Eq~\eqref{Jn_bound} holds for some $n$. Then by applying $T$ and using monotonicity, 
\begin{equation*}
T(\alpha \mathbf{1} + h + n \lambda \mathbf{1}) \leqslant T(J_n) \leqslant T(\beta \mathbf{1} + h + n \lambda \mathbf{1})
\end{equation*}
By translation invariance and $Th = \lambda \mathbf{1} + h$,
\begin{eqnarray*}
&T(\alpha \mathbf{1} + h + n \lambda \mathbf{1}) = \alpha \mathbf{1} + (n + 1) \lambda \mathbf{1} + h,\\
&T(\beta \mathbf{1} + h + n \lambda \mathbf{1}) = \beta \mathbf{1} + (n + 1) \lambda \mathbf{1} + h.
\end{eqnarray*}
Since $J_{n+1} = T J_n$, we derive Eq~\eqref{Jn_bound} for $n + 1$. Thus the bounds hold for all $n$. 
Fix any $i \in \X_0$. From Eq~\eqref{Jn_bound}, $\alpha + h(i) + n \lambda \leqslant J_n(i) \leqslant \beta + h(i) + n \lambda$, and dividing by $n$ gives
\begin{equation*}
\lambda + \frac{\alpha + h(i)}{n} \leqslant \frac{1}{n} J_n(i) \leqslant \lambda + \frac{\beta + h(i)}{n}.
\end{equation*}
Letting $n \rightarrow \infty$ yields $\lim_{n \rightarrow \infty} \frac{1}{n} J_n(i) = \lambda.$
By the definition of $\lambda^{\star}$, we conclude that $\lambda^{\star} = \lambda$.
\end{proof}
\noindent{\em Proof of Theorem~\ref{eigen_relation}:}
By definition of $(\Lambda, H)$, the pairs $(\lambda, h)$ satisfy $Th = \lambda \mathbf{1} + h$. Since $T$ is translation invariant, iteration this relation yields $T^n h_\ell = h_\ell + n \lambda_\ell \mathbf{1}$ for $\ell = 1, 2$ and $n \geqslant 0$. 
Moreover, by the nonexpansiveness of $T$ in the supremum norm, $\|T^n h_1 - T^n h_2\|_\infty \leqslant \|h_1 - h_2\|_\infty$ for all $n \geqslant 0$. 
Substituting the above expression for $T_n h_\ell$ gives 
\begin{equation*}
\|h_1 - h_2 + n (\lambda_1 - \lambda_2) \mathbf{1}\|_\infty = \|T^n h_1 - T^n h_2\|_\infty \leqslant \|h_1 - h_2\|_\infty.
\end{equation*}
If $\lambda_1 \neq \lambda_2$, then the left-hand side grows linearly in $n$ and diverge to $+\infty$, which contradicts uniform bound on the right-hand side. Hence $\lambda_1 = \lambda_2$. 
By Lemma~\ref{lambda_equal}, any eigenvalue $\lambda$ of $T$ must equal $\lambda^\star$. Thus $\lambda_1 = \lambda_2 = \lambda^{\star}$ and $\Lambda_1 = \Lambda_2 = \Lambda^\star$. 

We now prove that the associated eigenvectors are unique up to a scalar factor. Since $H_\ell = \exp(h_\ell)$ and $\lambda_\ell = \lambda^\star$ for $\ell = 1, 2$, using Definition~\ref{def:Th}, both $(\Lambda^\star, H_1)$ and $(\Lambda^\star, H_2)$ satisfy $H(i) = \max_v \min_u \frac{\exp(c_{uv}(i))}{\Lambda^\star} \sum_{j \in \X_0} P_{uv}^{\star}(j|i) H(j)$ for $i \in \X_0$. 
For each $i \in \X_0$, choose $v^{\star}(i) \in \arg\max_{v}\min_{u} \frac{\exp(c_{uv}(i))}{\Lambda^\star} \sum_{j \in \X_0} P_{uv}^{\star}(j|i)H_2(j)$. Then,
\begin{equation*}
H_2(i) = \min_{u} \frac{\exp(c_{uv^{\star}(i)}(i))}{\Lambda^\star} \sum_{j \in \X_0} P_{uv^{\star}(i)}^{\star}(j|i)H_2(j).
\end{equation*}
Next, for this fixed $v^{\star}(i)$, choose $u^{\prime}(i) \in \arg\min_u \frac{\exp(c_{uv^{\star}(i)}(i))}{\Lambda^\star} \sum_{j \in \X_0} P_{uv^{\star}(i)}^{\star}(j|i)H_1(j)$. Then, 
\begin{eqnarray*}
H_1(i) &= \max_{v}\min_{u} \frac{\exp(c_{uv}(i))}{\Lambda^\star} \sum_{j \in \X_0} P_{uv}^{\star}(j|i) H_1(j) \nn
\end{eqnarray*}
\begin{eqnarray*}
&\geqslant \min_u \frac{\exp(c_{uv^{\star}(i)}(i))}{\Lambda^\star} \sum_{j \in \X_0} P_{uv^{\star}(i)}^{\star}(j|i) H_1(j)\\
&\geqslant \frac{\exp(c_{u^{\prime}(i)v^{\star}(i)}(i))}{\Lambda^\star} \sum_{j \in \X_0} P_{u^{\prime}(i)v^{\star}(i)}^{\star}(j|i) H_1(j).
\end{eqnarray*}
From the expression for $H_2(i)$, for any choice of $u(i)$, 
\begin{equation*}
H_2(i) \leqslant \frac{\exp(c_{u(i)v^{\star}(i)}(i))}{\Lambda^\star} \sum_{j \in \X_0} P_{u(i)v^{\star}(i)}^{\star}(j|i)H_2(j).
\end{equation*}
Taking $u(i) = u^{\prime}(i)$ and dividing the two inequalities, 
\begin{equation*}
\frac{H_1(i)}{H_2(i)} \geqslant \frac{\sum_{j \in \X_0} P_{u^{\prime}(i)v^{\star}(i)}^{\star}(j|i)H_1(j)}{\sum_{j \in \X_0} P_{u^{\prime}(i)v^{\star}(i)}^{\star}(j|i)H_2(j)}.
\end{equation*}
Define the ratio $A(i) := \frac{H_1(i)}{H_2(i)}$ for $i \in \X_0$. Since $H_1$ and $H_2$ are strictly positive and $\X_0$ is finite, $A$ is bounded and strictly positive. Using $H_1(j) = A(j) H_2(j)$ in the inequality above 
\begin{align*}
A(i)& \geqslant \sum_{j \in \X_0} \frac{P_{u^{\prime}(i)v^{\star}(i)}^{\star}(j|i)H_2(j)}{\sum_{k \in \X_0} P_{u^{\prime}(i)v^{\star}(i)}^{\star}(k|i)H_2(k)} A(j)\nn
\end{align*}
\begin{align*}
&= \sum_{j \in \X_0} \bar{P}(j|i) A(j) \geqslant 0,
\end{align*}
where $\bar{P}(j|i) := \frac{P_{u^{\prime}(i)v^{\star}(i)}^{\star}(j|i)H_2(j)}{\sum_{k \in \X_0} P_{u^{\prime}(i)v^{\star}(i)}^{\star}(k|i)H_2(k)}$. The kernel $\bar{P}(j|i)$ is a stochastic matrix on $\X_0$ with the same support as $P_{u^{\prime}(i)v^{\star}(i)}^{\star}(\cdot|i)$, and by Assumption~\ref{assume:irr} it is irreducible. 

The rest of the proof follows technique from Lemma~2.1 in \cite{borkar2002q}.
Let $\{W_n\}_{n \geqslant 0}$ be the Markov chain on $\X_0$ with transition probabilities $\bar{P}(\cdot|\cdot)$. Then $A(i) \geqslant \sum_j \bar{P}(j|i) A(j) = \bE[A(W_{n + 1})|W_n = i].$
Thus $\{A(W_n)\}_{n \geqslant 0}$ is a bounded, nonnegative supermartingale and hence converges almost surely. 

Since $\bar{P}$ is irreducible on the finite state space $\X_0$, the chain $\{W_n\}$ is recurrent and visits every state infinitely often with probability one. If $A$ is not a constant vector on $\X_0$, the sequence $A(W_n)$ would take at least two distinct values infinitely often and thus could not converge almost surely. Therefore $A$ must be a constant vector on $\X_0$, i.e., there exists $c > 0$ such that $A(i) = c$ for all $i \in \X_0$. Thus, $H_1 = c H_2$. 
\QED

Thus, the positive action-value fixed point utilized below is well defined. If $H$ is any positive eigenvector corresponding to $\Lambda^\star$, set $Q^\star(i, u, v) = \frac{\exp(c_{uv}(i))}{\Lambda^\star} 
\sum_{j \in \X_0} P_{uv}^\star(j|i)H (j).$

Then $\max_{v \in \V} \min_{u \in \U} Q^\star(i, u, v) = H(i)$, hence $F(Q^\star) = Q^\star$. Conversely, any strictly positive fixed point of $F$ yields $H(i) = \max_{v \in \V} \min_{u \in \U} Q(i,u,v)$, which solves the multiplicative Bellman-Perron equation. Thus, Theorem~\ref{eigen_relation} implies that $Q^\star$ is unique up to multiplication by a positive scalar. We use normalized representative $Q^\star(i_0, u_0, v_0) = \Lambda^\star$.
\subsection{Proving global asymptotical stability of $Q^\star$}
\begin{lemma}\label{support_lemma1_secondary}
For every $(i,u,v)$, we have
\begin{align*}
&\min_{\{(j,u^{\prime},v^{\prime}): P_{uv}^{\star}(j|i) > 0\}} \frac{q(j,u^{\prime},v^{\prime})}{Q^{\star}(j,u^{\prime},v^{\prime})} \leqslant \frac{F_{iuv}(q)}{Q^{\star}(i,u,v)}\\
&= \frac{F_{iuv}(q)}{F_{iuv}(Q^\star)} \leqslant \max_{\{(j,u^{\prime},v^{\prime}): P_{uv}^{\star}(j|i) > 0\}} \frac{q(j,u^{\prime},v^{\prime})}{Q^{\star}(j,u^{\prime},v^{\prime})}.
\end{align*}
\end{lemma}
\begin{proof}
Since $Q^\star$ is a positive fixed point of $F$, we have
\begin{align*}
Q^{\star}(i,u,v) &= F_{iuv}(Q^\star)\\
&= \frac{\exp(c_{uv}(i))}{\Lambda^\star} \sum_{j \in \X_0} P_{uv}^{\star}(j|i) \max_{v^{\prime}} \min_{u^{\prime}} Q^{\star}(j,u^{\prime},v^{\prime}).
\end{align*}
The multiplicate constant $\frac{\exp(c_{uv}(i))}{\Lambda^\star}$ cancels in the ratio, so
\begin{equation*}
\frac{F_{iuv}(q)}{Q^\star(i,u,v)} = \frac{F_{iuv}(q)}{F_{iuv}(Q^\star)} = \frac{\sum_{j} P_{uv}^\star(j|i) A_j}{\sum_{j} P_{uv}^\star(j|i) B_j},
\end{equation*}
where for each $j \in \X_0$, $A_j := \max_{v^{\prime}} \min_{u^{\prime}} q(j,u^{\prime},v^{\prime})$ and $B_j := \max_{v^{\prime}} \min_{u^{\prime}} Q^{\star}(j,u^{\prime},v^{\prime})$. 
Fix $j$ with $P_{uv}^{\star}(j|i) > 0$. Define the point-wise ratio $R(j,u^{\prime},v^{\prime}) := \frac{q(j,u^{\prime},v^{\prime})}{Q^{\star}(j,u^{\prime},v^{\prime})}$, and set $m_j := \min_{u^{\prime},v^{\prime}} R(j,u^{\prime},v^{\prime})$, $M_j := \max_{u^{\prime},v^{\prime}} R(j,u^{\prime},v^{\prime})$. Then for all $(u^{\prime},v^{\prime})$, $m_j \leqslant R(j,u^{\prime},v^{\prime}) \leqslant M_j$. Thus $m_j Q^{\star}(j,u^{\prime},v^{\prime}) \leqslant q(j,u^{\prime},v^{\prime}) \leqslant M_j Q^{\star}(j,u^{\prime},v^{\prime})$. 

Fix $v^{\prime}$. Taking the minimum over $u^{\prime}$ yields 
\begin{equation*}
m_j \min_{u^{\prime}} Q^{\star}(j,u^{\prime},v^{\prime}) \leqslant \min_{u^{\prime}} q(j,u^{\prime},v^{\prime}) \leqslant M_j \min_{u^{\prime}} Q^{\star}(j,u^{\prime},v^{\prime}).
\end{equation*}
Now take the maximum over $v^{\prime}$, we have 
\begin{align*}
m_j \max_{v^{\prime}} \min_{u^{\prime}} Q^{\star}(j,u^{\prime},v^{\prime}) &\leqslant \max_{v^{\prime}} \min_{u^{\prime}} q(j,u^{\prime},v^{\prime})\\
&\leqslant M_j \max_{v^{\prime}} \min_{u^{\prime}} Q^{\star}(j,u^{\prime},v^{\prime}).
\end{align*}
Thus, $m_j B_j \leqslant A_j \leqslant M_j B_j$, thus $m_j \leqslant \frac{A_j}{B_j} \leqslant M_j$. 
Define 
\begin{equation*}
m := \min_{\{j: P_{uv}^{\star}(j|i) > 0\}} m_j = \min_{\{(j,u^{\prime},v^{\prime}): P_{uv}^{\star}(j|i) > 0\}} \frac{q(j,u^{\prime},v^{\prime})}{Q^{\star}(j,u^{\prime},v^{\prime})},
\end{equation*}
\begin{equation*}
M := \max_{\{j: P_{uv}^{\star}(j|i) > 0\}} M_j = \max_{\{(j,u^{\prime},v^{\prime}): P_{uv}^{\star}(j|i) > 0\}} \frac{q(j,u^{\prime},v^{\prime})}{Q^{\star}(j,u^{\prime},v^{\prime})}.
\end{equation*}
Then for every $j$ with $P_{uv}^{\star}(j|i) > 0$, $m \leqslant m_j \leqslant \frac{A_j}{B_j} \leqslant M_j \leqslant M$. 
Finally, we write the ratio as
\begin{equation*}
\frac{F_{iuv}(q)}{Q^{\star}(i,u,v)} = \frac{\sum_j P_{uv}^{\star}(j|i) B_j \cdot \frac{A_j}{B_j}}{\sum_j P_{uv}^{\star}(j|i) B_j} = \sum_j w_j \frac{A_j}{B_j},
\end{equation*}
where $w_j := \frac{P_{uv}^{\star}(j|i) B_j}{\sum_k P_{uv}^{\star}(k|i) B_k}$ are nonnegative weights that sum to $1$. Since each $\frac{A_j}{B_j} \in [m,M]$, this convex combination also lies in $[m,M]$. Thus $m \leqslant \frac{F_{iuv}(q)}{Q^{\star}(i,u,v)} \leqslant M$.
\end{proof}

\begin{corollary}\label{support_corollary1_secondary}
$(q)_m \leqslant (F(q))_m \leqslant (F(q))_M \leqslant (q)_M$.
\end{corollary}
\begin{proof}
The claim follows immediately from Lemma~\ref{support_lemma1_secondary} by taking the minimum and maximum over all $(i,u,v)$ of the point-wise inequalities.
\end{proof}

\begin{lemma}\label{support_lemma2_secondary}
Let $Q(t)$ be the positive solution of the  ODE $\dot{Q}(t) = F(Q(t)) - Q(t),$
for $t \geqslant 0$, $Q(0) > 0$. Then $(Q(t))_M$ is non-increasing, and $(Q(t))_m$ is non-decreasing.
\end{lemma}
\begin{proof}
Fix $N \geqslant 1$ and set $\epsilon := 2^{-N}$. Consider the explicit Euler discretization of the ODE, 
\begin{equation*}
Q^{(\epsilon)}((n + 1) \epsilon) = Q^{(\epsilon)}(n \epsilon) + \epsilon (F(Q^{(\epsilon)}(n \epsilon)) - Q^{(\epsilon)}(n \epsilon)),
\end{equation*}
for $n = 0, 1, 2, \ldots$, with initial condition $Q^{(\epsilon)}(0) = Q(0) > 0$. $Q_n := Q^{(\epsilon)}(n \epsilon)$ for $n = 0, 1, 2, \cdots$. Then for each $(i, u, v)$
\begin{equation*}
Q_{n + 1}(i, u, v) = (1 - \epsilon) Q_n (i, u, v) + \epsilon F_{iuv}(Q_n).
\end{equation*}
Dividing by $Q^{\star}(i, u, v)$ and minimizing over $(i, u, v)$ yields
\begin{align*}
(Q_{n + 1})_m &= \min_{i, u, v} \frac{Q_{n + 1}(i ,u, v)}{Q^{\star}(i, u, v)} \\
&\geqslant (1 - \epsilon) \min_{i, u, v} \frac{Q_n(i, u, v)}{Q^{\star}(i, u, v)} + \epsilon \min_{i, u, v} \frac{F_{iuv}(Q_n)}{Q^{\star}(i, u, v)} \\
&= (1 - \epsilon) (Q_n)_m + \epsilon (F(Q_n))_m,
\end{align*}
where we used that for any two families $(a_n)$, $(b_n)$ and $\theta \in [0, 1]$, $\min_n \Big[(1 - \theta) a_n + \theta b_n\Big] \geqslant (1 - \theta) \min_n a_n + \theta \min_n b_n$. By Corollary~\ref{support_corollary1_secondary}, we have
$(F(Q_n))_m \geqslant (Q_n)_m.$
Hence
\begin{equation*}
(Q_{n + 1})_m \geqslant (1 - \epsilon) (Q_n)_m + \epsilon (Q_n)_m = (Q_n)_m.
\end{equation*}
Thus, for each fixed step size $\epsilon$, the sequence $\{(Q_n)_m\}_{n \geqslant 0}$ is non-decreasing in $n$.
An analogous argument, starting from
\begin{equation*}
Q_{n + 1}(i, u, v) = (1 - \epsilon) Q_n(i, u, v) + \epsilon F_{iuv}(Q_n),
\end{equation*}
and using $\max_n \Big[(1 - \theta) a_n + \theta b_n\Big] \leqslant (1 - \theta) \max_n a_n + \theta \max_n b_n,$
together with $(F(q))_M \leqslant (q)_M$, shows that the sequence $\{(Q_n)_M\}_{n \geqslant 0}$ is non-increasing in $n$.
Now, for each $\epsilon$ we extend $\{Q_n\}$ to a continuous-time process $Q^{(\epsilon)}(t)$ by linear interpolation on each interval $[n \epsilon, (n + 1) \epsilon]$. Since $F$ is Lipschitz on the order interval determined by $Q(0)$ and $Q^{\star}$, the Euler scheme converges uniformly on bounded time intervals to the unique solution $Q(t)$ of the ODE $\dot{Q} = F(Q) - Q$.
Fix $0 \leqslant s < t$ and choose dyadic rationals $s_{\epsilon}, t_{\epsilon}$ such that $s_{\epsilon} \rightarrow s$, $t_{\epsilon} \rightarrow t$, and $s_{\epsilon} < t_{\epsilon}$. For each $\epsilon$ we have monotonicity, 
\begin{equation*}
(Q^{(\epsilon)}(s_{\epsilon}))_m \leqslant (Q^{(\epsilon)}(t_{\epsilon}))_m.
\end{equation*}
By the uniform convergence $Q^{(\epsilon)} \rightarrow Q$ on compact intervals, together with $s_{\epsilon} \rightarrow s$ and $t_{\epsilon} \rightarrow t$, we have $Q^{(\epsilon)}(s_{\epsilon}) \rightarrow Q(s)$ and $Q^{(\epsilon)}(t_{\epsilon}) \rightarrow Q(t)$. Since the map $q \rightarrow (q)_m$ is continuous in finite dimension, it follows that
\begin{eqnarray*}
&(Q^{(\epsilon)}(s_{\epsilon}))_m \leqslant (Q^{(\epsilon)}(t_{\epsilon}))_m \text{~and~}(Q(s))_m \leqslant (Q(t))_m.
\end{eqnarray*}
Thus $(Q(t))_m$ is non-decreasing on $[0, \infty]$.
The same argument, applied to the non-increasing sequence $\{(Q_n)_M\}_{n \geqslant 0}$, shows that $(Q(t))_M$ is non-increasing.
\end{proof}

\begin{corollary}\label{support_corollary2_secondary}
For $t > 0$, $(Q(t))_m \geqslant (Q(0))_m > 0.$
\end{corollary}
\begin{proof}
By Lemma~\ref{support_lemma2_secondary}, $(Q(t))_m := \min_{i, u, v} \frac{Q(t, i, u, v)}{Q^{\star}(i, u, v)}$ is non-decreasing in $t$. In particular, for every $t > 0$, $(Q(t))_m \geqslant (Q(0))_m$. Since $Q(0) > 0$ and $Q^{\star} > 0$ componentwise, 
$
(Q(0))_m = \min_{i, u, v} \frac{Q(0, i, u, v)}{Q^{\star}(i, u, v)} > 0.
$
\end{proof}

\begin{theorem}\label{Risk:Thm2}
Let $Q(t)$ be the positive solution of the  ODE
\begin{equation}\label{ODE_equation}
\dot{Q}(t) = F(Q(t)) - Q(t),
\end{equation}
for $t \geqslant 0$ with initial condition $Q(0) > 0$. Then every such solution $Q(\cdot)$ converges to $\mathrm{H} = \{r Q^{\star}: r > 0\}$, where $Q^{\star}$ is the strictly positive fixed point of $F$, i.e., $Q^{\star} = F(Q^{\star})$. 
\end{theorem}
\begin{proof}
Let $Q(\cdot)$ be a solution of Eq.~\eqref{ODE_equation} with $Q(0) > 0$. Define the $\omega$-limit set of this trajectory by 
\begin{equation*}
\Omega := \big\{x: \exists~ t(n) \rightarrow \infty \quad \text{such that} \quad Q(t(n)) \rightarrow x\big\}. 
\end{equation*}
By Lemma~\ref{support_lemma2_secondary}, $(Q(t))_m$ is non-decreasing and $(Q(t))_M$ is non-increasing in $t$, and Corollary~\ref{support_corollary2_secondary} ensures $(Q(t))_m \geqslant (Q(0))_m > 0$ for all $t > 0$. Hence $Q(\cdot)$ is bounded. It follows that $\Omega$ is nonempty, compact, and invariant under Eq.~\eqref{ODE_equation}. 
Since $(Q(t))_m$ and $(Q(t))_M$ are monotone and bounded,
\begin{equation*}
\beta_1 := \lim_{t \rightarrow \infty} (Q(t))_m, \quad \beta_2 := \lim_{t \rightarrow \infty} (Q(t))_M
\end{equation*}
exist and satisfy $0 < \beta_1 \leqslant \beta_2 < \infty$. Define
\begin{equation*}
B := \Big\{q > 0: (q)_m = \beta_1, (q)_M = \beta_2\Big\}. 
\end{equation*}
By continuity of the maps $q \mapsto (q)_m$ and $q \mapsto (q)_M$ in finite dimension, we have $\Omega \subset B$. Moreover, for every $q \in B$, 
\begin{equation*}
\beta_1 Q^{\star}(i, u, v) \leqslant q(i, u, v) \leqslant \beta_2 Q^{\star}(i, u, v),
\end{equation*}
and by definition of $(q)_m$ and $(q)_M$, there is at least one component where equality holds with $\beta_1$ and at least one where equality holds with $\beta_2$. 
For each state-action tuple $(m, n, c)$, define the closed subset 
\begin{equation*}
D_{mnc} := \Big\{q \in B: q(m, n, c) = \beta_2 Q^{\star}(m, n, c)\Big\}.
\end{equation*}
Every $q \in B$ attains its maximal ratio $(q)_M = \beta_2$ at some component $(m, n, c)$, so $B = \bigcup_{(m, n, c)} D_{mnc}.$
Consequently, $\Omega \subset B \subset \bigcup_{(m, n, c)} D_{mnc},$
and therefore there exists at least one tuple $(m, n, c)$ such that 
$\Omega \cap D_{mnc} \neq \emptyset.$

Fix such a tuple and denote $D := D_{mnc}$. 
Now pick $x_0 \in \Omega \cap D$ and let $\widetilde{Q}(\cdot)$ be the solution of Eq.~\eqref{ODE_equation} with initial condition $\widetilde{Q}(0) = x_0$. Since $\Omega$ is invariant, we have $\widetilde{Q}(t) \in \Omega \subset B$ for all $t \geqslant 0$. Thus the continuous trajectory $\widetilde{Q}: [0, \infty) \rightarrow B$ takes values in $B$, and for each $t$ we have $\widetilde{Q}(t) \in D_{mnc^{\prime}}$ for at least one $(m, n, c^{\prime})$. 
Because there are only finitely many indices $(m, n, c)$, the sets $D_{mnc}$ are finitely many closed subsets of $B$. Since $\widetilde{Q}(\cdot)$ is continuous, 
\begin{equation*}
T_{mnc} := \{t \geqslant 0: \widetilde{Q}(t) \in D_{mnc}\}
\end{equation*}
are closed subsets of $[0, \infty)$, and $[0, \infty) = \bigcup_{(m, n, c)} T_{mnc}.$
Given that the finitely many closed sets $T_{mnc}$ cover $[0, \infty)$, it follows that at least one of these sets must have nonempty interior; otherwise their union would show empty interior, which contradicts the fact that $[0, \infty)$ has nonempty interior. Thus, there exists a tuple $(\bar{m}, \bar{n}, \bar{c})$ and an interval$(a, b) \subset T_{\bar{m}\bar{n}\bar{c}}$ with $0 \leqslant a < b < \infty$.
By time-translation invariance of ODE~\eqref{ODE_equation}, we may without loss of generality assume $a = 0$ by redefining the initial time. Thus, for some tuple $(m, n, c)$, 
\begin{equation*}
\widetilde{Q}(t) \in D_{mnc},
\end{equation*}
for all $t \in [0, \tau]$ and some $\tau > 0$. By the definition of $D_{mnc}$, 
\begin{equation*}
\widetilde{Q}_{mnc}(t) = \beta_2 Q^{\star}(m, n, c)
\end{equation*}
for $t \in [0, \tau]$. Since $\widetilde{Q}$ is differentiable, this implies 
\begin{equation*}
\frac{d}{dt} \widetilde{Q}_{mnc}(t) = 0
\end{equation*}
for $t \in [0, \tau]$. From the ODE~\eqref{ODE_equation}, we thus obtain
\begin{equation*}
F_{mnc}(\widetilde{Q}(t)) = \widetilde{Q}_{mnc}(t) = \beta_2 Q^{\star}(m, n, c)
\end{equation*}
for $t \in [0, \tau]$. Dividing by $Q^{\star}(m, n, c) = F_{mnc}(Q^{\star})$, we get
\begin{equation*}
\beta_2 = \frac{\widetilde{Q}_{mnc}(t)}{Q^{\star}(m, n, c)} = \frac{F_{mnc}(\widetilde{Q}(t))}{F_{mnc}(Q^{\star})}.
\end{equation*}
Next, writing out $F_{mnc}$ explicitly, we have
\begin{align*}
&\beta_2= \sum_{j \in \X_0} w_{mnc}(j) \frac{\max_{v^{\prime}} \min_{u^{\prime}} \widetilde{Q}(j,u^{\prime},v^{\prime})}{\max_{v^{\prime}} \min_{u^{\prime}} Q^{\star}(j,u^{\prime},v^{\prime})},
\end{align*}
where $w_{mnc}(j) := \frac{P_{nc}^{\star}(j|m) \max_{v^{\prime}} \min_{u^{\prime}} Q^{\star}(j,u^{\prime},v^{\prime})}{\sum_{k \in \X_0} P_{nc}^{\star}(k|m) \max_{v^{\prime}} \min_{u^{\prime}} Q^{\star}(k,u^{\prime},v^{\prime})}$ are nonnegative weights with $\sum_j w_{mnc}(j) = 1$. 
Since $\widetilde{Q}(t) \in B$ for all $t$, in particular at our fixed $t \in [0, \tau]$ we have 
\begin{equation*}
\beta_1 Q^{\star}(j,u^{\prime},v^{\prime}) \leqslant \widetilde{Q}(j,u^{\prime},v^{\prime}) \leqslant \beta_2 Q^{\star}(j,u^{\prime},v^{\prime}) \quad \forall (j,u^{\prime},v^{\prime}).
\end{equation*}
Therefore, for each $j$, $\beta_1 \leqslant \frac{\max_{v^{\prime}} \min_{u^{\prime}} \widetilde{Q}(j,u^{\prime},v^{\prime})}{\max_{v^{\prime}} \min_{u^{\prime}} Q^{\star}(j,u^{\prime},v^{\prime})} \leqslant \beta_2.$
We have thus written $\beta_2$ as a convex combination of numbers lying in the interval $[\beta_1, \beta_2]$, $\beta_2 = \sum_{j \in \X_0} w_{mnc}(j) \rho_j,$
where $\rho_j := \frac{\max_{v^{\prime}} \min_{u^{\prime}} \widetilde{Q}(j,u^{\prime},v^{\prime})}{\max_{v^{\prime}} \min_{u^{\prime}} Q^{\star}(j,u^{\prime},v^{\prime})}$. 
This is possible only if $\rho_j = \beta_2$ for all $j$ with $w_{mnc}(j) > 0$. Given that $\max_{v'} \min_{u'} Q^\star(j, u', v') > 0$, 
\begin{equation*}
\frac{\max_{v'} \min_{u'} \widetilde{Q}_{ju'v'}(t)}
{\max_{v'} \min_{u'} Q^\star(j, u', v')} = \beta_2
\end{equation*}
for every $j \in \X_0$ such that $P_{nc}^\star(j|m) > 0$. As the preceding argument is valid for each fixed $t \in [0, \tau]$, the same conclusion holds across the entire interval $[0, \tau]$. We will now propagate this equality. Suppose that, for some state $i$, the given equality is valid across a nondegenerate subinterval of $[0, \tau]$. Due to the finite nature of $\V$ and continuity of $\widetilde{Q}(\cdot)$, it is possible to restrict to a smaller nondegenerate subinterval where a specific fixed $v$ achieves the maximum in $\max_{v'} \min_{u'} \widetilde{Q}_{iu'v'}(t)$ throughout that subinterval. The componentwise bound $\widetilde{Q}_{iuv}(t) \leqslant \beta_2 Q^\star(i, u, v)
$
then requires this same $v$ to achieve the maximum in $\max_{v'} \min_{u'} Q^\star(i, u', v').$
Furthermore, for $u$ that minimizes $\min_{u'} Q^\star(i, u', v)$, $\widetilde{Q}_{iuv}(t) = \beta_2 Q^\star(i, u, v)$
throughout that subinterval. For this fixed component, the derivative is zero on the subinterval. Thus, by the ODE, $F_{iuv}(\widetilde{Q}(t)) = \beta_2 Q^\star(i, u, v) = \beta_2 F_{iuv}(Q^\star).$
By dividing by $F_{iuv}(Q^\star)$ and applying the same convex-combination argument, we obtain
\begin{equation*}
\frac{\max_{v'} \min_{u'} \widetilde{Q}_{ju'v'}(t)}
{\max_{v'} \min_{u'} Q^\star(j, u', v')} = \beta_2
\end{equation*}
for all $j \in \X_0$ with $P_{uv}^\star(j|i) > 0$. By repeating this finite propagation argument, the propagation cannot stop before reaching all states in $\X_0$. If it stopped on a proper nonempty set of states, then the fixed action pairs selected on those states, completed arbitrarily outside that set, would make that set closed under the corresponding restricted kernel. Given that $\bP_{uv}^0(j|i) = d_{uv}(i) P_{uv}^\star(j|i)$ and $d_{uv}(i) > 0$, the kernels $\bP_{uv}^0$ and $P_{uv}^\star$ have the same positive entries. This contradicts Assumption~\ref{assume:irr} on irreducibility over $\X_0$. Thus, after restricting to a nondegenerate interval and relabeling it as $[0, \tau]$,
\begin{equation*}
\frac{\max_{v'} \min_{u'} \widetilde{Q}_{ju'v'}(t)}
{\max_{v'} \min_{u'} Q^\star(j, u', v')} = \beta_2, ~\text {for all~} j \in \X_0,~ t \in [0, \tau].
\end{equation*}
Consequently, for all $(i, u, v)$ and all $t \in [0, \tau]$, 
\begin{align*}
F_{iuv}(\widetilde{Q}(t)) &= \frac{\exp(c_{uv}(i))}{\Lambda^\star} \sum_{j \in \X_0} P_{uv}^{\star}(j|i) \max_{v^{\prime}} \min_{u^{\prime}} \widetilde{Q}(j,u^{\prime},v^{\prime}) \\
&= \beta_2 F_{iuv}(Q^{\star}) = \beta_2 Q^{\star}(i,u,v).
\end{align*}
Hence, along $[0, \tau]$, the ODE~\eqref{ODE_equation} for $\widetilde{Q}$ reads
\begin{equation*}
\frac{d}{dt} \widetilde{Q}_{iuv}(t) = F_{iuv}(\widetilde{Q}(t)) - \widetilde{Q}_{iuv}(t) = \beta_2 Q^{\star}(i,u,v) - \widetilde{Q}_{iuv}(t).
\end{equation*}
By dividing this equation by $Q^\star(i, u, v)$, for every $(i, u, v)$,
\begin{eqnarray*}
&\frac{d}{dt} \frac{\widetilde{Q}_{iuv}(t)}{Q^\star(i, u, v)} = \beta_2 - \frac{\widetilde{Q}_{iuv}(t)}{Q^\star(i, u, v)}, \quad t \in [0, \tau],\\
&\frac{\widetilde{Q}_{iuv}(t)}{Q^\star(i, u, v)} = \beta_2 + e^{-t} \left(\frac{\widetilde{Q}_{iuv}(0)}{Q^\star(i, u, v)} - \beta_2\right), \quad t \in [0, \tau].
\end{eqnarray*}
If $\beta_1 < \beta_2$, then every component whose initial ratio equals $\beta_1$ becomes strictly larger than $\beta_1$ for every positive $t \in [0, \tau]$. Every component whose initial ratio is already larger than $\beta_1$ will continue to be larger than $\beta_1$ for all sufficiently small positive values of $t$. Given the finite number of components, it follows that the minimum componentwise ratio of $\widetilde{Q}(t)$ is strictly larger than $\beta_1$ for all sufficiently small positive values of $t$, thereby contradicting the condition that $\widetilde{Q}(t) \in B$. Thus, $\beta_1 = \beta_2 =: \beta > 0$. 
Finally, for each $(i, u, v)$ and all $t$, 
\begin{equation*}
(Q(t))_m \leqslant \frac{Q(t, i, u, v)}{Q^{\star}(i, u, v)} \leqslant (Q(t))_M,
\end{equation*}
and both endpoints $(Q(t))_m$ and $(Q(t))_M$ converge to $\beta$. Therefore for all $(i, u, v)$, $\frac{Q(t, i, u, v)}{Q^{\star}(i, u, v)} \rightarrow \beta$
which implies $Q(t) \rightarrow \beta Q^{\star} \in H$
as $t \rightarrow \infty$.
\end{proof}

\begin{lemma}\label{support_lemma3_secondary}
Let $Q(t)$ be the positive solution of the ODE
$\dot{Q}(t) = G(Q(t)) - Q(t),$
for $t \geqslant 0$ with initial condition $Q(0) > 0$. Then $Q(t) \rightarrow Q^{\star}$ as $t \rightarrow \infty$. In particular, for the tuple $(i_0, u_0, v_0)$, we have $Q_{i_{0}u_{0}v_{0}}(t) \rightarrow \Lambda^{\star} = e^{\lambda^{\star}}$.
\end{lemma}
\begin{proof}
The proof follows Lemma~4.1 of \cite{borkar2002q}.
\end{proof}

\begin{lemma}\label{support_lemma4_secondary}
For every $\epsilon > 0$, there exists $\delta > 0$ such that when $\|Q(0) - Q^{\star}\| < \delta$, the corresponding solution of
$\dot{Q}(t) = G(Q(t)) - Q(t)$
satisfies
$|Q_{i_{0}u_{0}v_{0}}(t) - \Lambda^{\star}| \leqslant \epsilon$, $t \geqslant 0$.
\end{lemma}
\begin{proof}
The proof follows Lemma~4.2 of \cite{borkar2002q}.
\end{proof}
\noindent{\em Proof of Theorem~\ref{Risk:Thm3}:}
Define $\Phi(q) := \log\frac{(q)_M}{(q)_m} \geqslant 0.$
Then, $\Phi(q) = 0$ if and only if $(q)_M = (q)_m$, which is equivalent to $q$ being a positive scalar multiple of $Q^{\star}$. For $\alpha > 0$, 
\begin{equation*}
(\alpha q)_m = \alpha (q)_m, \quad (\alpha q)_M = \alpha (q)_M,
\end{equation*}
and hence $\Phi(\alpha q) = \Phi(q),$
so $\Phi$ is homogeneous of degree zero. 
Recall that $Q(t)$ denotes the positive solution of $\dot{Q}(t) = G(Q(t)) - Q(t),$
while $\bar{Q}(t)$ denotes the positive solution of $\dot{\bar{Q}}(t) = F(\bar{Q}(t)) - \bar{Q}(t)$,
both starting from the same initial condition $Q(0) = \bar{Q}(0) > 0$. By definition of $G$, we have
\begin{equation*}
G(Q) = \frac{\Lambda^{\star}}{Q(i_{0}u_{0}v_{0})} F(Q),
\end{equation*}
and therefore the ODE for $Q(\cdot)$ can be written as
\begin{equation*}
\dot{Q}(t) = \frac{\Lambda^{\star}}{Q_{i_{0}u_{0}v_{0}}(t)} F(Q(t)) - Q(t).
\end{equation*}

Define
$\gamma(t) := \frac{Q_{i_{0}u_{0}v_{0}}(t)}{\Lambda^{\star}} > 0,
\tau(t) := \int_0^t \frac{1}{\gamma(s)} ds = \int_0^t \frac{\Lambda^{\star}}{Q_{i_{0}u_{0}v_{0}}(s)} ds.
$
Since $G(Q) > 0$, we have
\begin{equation*}
\dot{Q}_{i_{0}u_{0}v_{0}}(t) = G_{i_{0}u_{0}v_{0}}(Q(t)) - Q_{i_{0}u_{0}v_{0}}(t) \geqslant - Q_{i_{0}u_{0}v_{0}}(t),
\end{equation*}
which implies $Q_{i_{0}u_{0}v_{0}}(t) \geqslant e^{-t} Q_{i_{0}u_{0}v_{0}}(0) > 0$, 
for all $t \geqslant 0$. Hence $\gamma(t) > 0$ and $\tau(t)$ is well-defined and strictly increasing, so it allows an inverse $t = \tau^{-1}(s)$. 

Next, define $\phi(t)$ as the solution of $\frac{\dot{\phi}(t)}{\phi(t)} = \frac{1}{\gamma(t)} - 1 = \frac{\Lambda^{\star}}{Q_{i_{0}u_{0}v_{0}}(t)} - 1,$
with $\phi(0) = 1$. Integrating, we obtain $\phi(t)= \exp\Big(\int_0^t (\frac{\Lambda^{\star}}{Q_{i_{0}u_{0}v_{0}}(s)} - 1) ds\Big).
$
Define $Q^{\prime}(\cdot)$ as $Q^{\prime}(s) := \frac{Q(t)}{\phi(t)}$
with $t = \tau^{-1}(s)$. Differentiating and using the chain rule, $\frac{d}{ds} Q^{\prime}(s) = \bigg(\frac{1}{\phi(t)} \Big(\frac{1}{\gamma(t)} F(Q(t)) - Q(t)\Big) - \frac{\dot{\phi}(t)}{\phi(t)^2} Q(t)\bigg) \gamma(t).$
Since $Q^{\prime}(s) = \frac{Q(t)}{\phi(t)}$ and $F$ is positively homogeneous of degree one, we have $F(Q(t)) = F(\phi(t) Q^{\prime}(s)) = \phi(t) F(Q^{\prime}(s)).$
Substituting this into the above expression gives $\frac{d}{ds} Q^{\prime}(s)= F(Q^{\prime}(s)) - \Big(\gamma(t) + \gamma(t) \frac{\dot{\phi}(t)}{\phi(t)}\Big) Q^{\prime}(s).$
By definition of $\phi(t)$, $\gamma(t) + \gamma(t) \frac{\dot{\phi}(t)}{\phi(t)} = \gamma(t) + \gamma(t) \big(\frac{1}{\gamma(t)} - 1\big) = 1$. 
Thus, $\frac{d}{ds} Q^{\prime}(s) = F(Q^{\prime}(s)) - Q^{\prime}(s).$
The initial condition
\begin{equation*}
Q^{\prime}(0) = \frac{Q(\tau^{-1}(0))}{\phi(\tau^{-1}(0))} = \frac{Q(0)}{\phi(0)} = Q(0).
\end{equation*}
Hence $Q^{\prime}(\cdot)$ is the unique solution of the ODE $\dot{\bar{Q}}(s) = F(\bar{Q}(s)) - \bar{Q}(s)$
with initial condition $Q(0) = \bar{Q}(0)$, i.e., $Q^{\prime}(s) = \bar{Q}(s)$. Rewriting in terms of the original time $t$ yields
\begin{equation*}
Q(t) = \phi(t) Q^{\prime}(\tau(t)) = \phi(t) \bar{Q}(\tau(t)).
\end{equation*}

Using the homogeneity of $(\cdot)_m$ and $(\cdot)_M$, for every $t \geqslant 0$, 
\begin{align*}
(Q(t))_m &= (\phi(t) \bar{Q}(\tau(t)))_m = \phi(t) (\bar{Q}(\tau(t)))_m, \\
(Q(t))_M &= (\phi(t) \bar{Q}(\tau(t)))_M = \phi(t) (\bar{Q}(\tau(t)))_M,\\
\frac{(Q(t))_M}{(Q(t))_m} &= \frac{(\bar{Q}(\tau(t)))_M}{(\bar{Q}(\tau(t)))_m}~\text{and~} \Phi(Q(t)) = \Phi(\bar{Q}(\tau(t))).
\end{align*}

By Lemma~\ref{support_lemma2_secondary}, for the ODE $\dot{\bar{Q}} = F(\bar{Q}) - \bar{Q}$, $(\bar{Q}(t))_M$ is non-increasing and $(\bar{Q}(t))_m$ is non-decreasing. Therefore $\frac{(\bar{Q}(t))_M}{(\bar{Q}(t))_m}$ and $\Phi(\bar{Q}(t))$ are non-increasing. Since $\tau(t)$ is strictly increasing, it follows that $\Phi(Q(t)) = \Phi(\bar{Q}(\tau(t)))$ is also non-increasing. In particular, for all $t \geqslant 0$, $\Phi(Q(t)) \leqslant \Phi(Q(0)).$

Now we establish Lyapunov stability of $Q^{\star}$ for the ODE $\dot{Q} = G(Q) - Q$. By Lemma~\ref{support_lemma4_secondary}, for any $\epsilon > 0$, there exists $\delta_s > 0$ such that whenever $\|Q(0) - Q^{\star}\|_{\infty} < \delta_s$, 
\begin{equation*}
\big|\frac{Q_{i_{0}u_{0}v_{0}}(t)}{\Lambda^{\star}} - 1\big| < \epsilon_s
\end{equation*}
for all $t \geqslant 0$. Equivalently, for all $t \geqslant 0$, 
\begin{equation*}
\gamma(t) = \frac{Q_{i_{0}u_{0}v_{0}}(t)}{\Lambda^{\star}} \in(1 - \epsilon_s, 1 + \epsilon_s).
\end{equation*}

By the definition of $(Q(t))_m$ and $(Q(t))_M$, we have
\begin{equation*}
(Q(t))_m \leqslant \gamma(t) \leqslant (Q(t))_M.
\end{equation*}
Since $\Phi(Q(t)) \leqslant \Phi(Q(0))$ and $\Phi(Q(t)) = \log\frac{(Q(t))_M}{(Q(t))_m}$,
\begin{eqnarray*}
&\frac{(Q(t))_M}{(Q(t))_m} \leqslant \exp(\Phi(Q(0))) := e^{\sigma} \text{~and~}(Q(t))_M \leqslant e^{\sigma} (Q(t))_m.
\end{eqnarray*}
Combining these inequalities yields
\begin{equation*}
\gamma(t) \leqslant (Q(t))_M \leqslant e^{\sigma} (Q(t))_m~\text{and~} (Q(t))_m \geqslant \gamma(t) e^{-\sigma},
\end{equation*}
\begin{equation}
\gamma(t) e^{-\sigma} \leqslant (Q(t))_m \leqslant \gamma(t) \leqslant (Q(t))_M \leqslant \gamma(t) e^{\sigma},~\text{for~} t \geqslant 0. \label{eq:RiskThm3_basic_bounds}
\end{equation}
For any $(i, u, v)$ and any $t \geqslant 0$, we also have
\begin{equation*}
(Q(t))_m \leqslant \frac{Q_{iuv}(t)}{Q_{iuv}^{\star}} \leqslant (Q(t))_M.
\end{equation*}
By Eq.~\eqref{eq:RiskThm3_basic_bounds}, for all $(i, u, v)$ and $t \geqslant 0$,
\begin{equation*}
\gamma(t) e^{-\sigma} \leqslant \frac{Q_{iuv}(t)}{Q_{iuv}^{\star}} \leqslant \gamma(t) e^{\sigma}.
\end{equation*}

Let $Q_{\min}^{\star} := \min_{i,u,v} Q_{iuv}^{\star} > 0, \quad Q_{\max}^{\star} := \max_{i,u,v} Q_{iuv}^{\star}.$
Fix $\epsilon > 0$ and choose $\eta \in (0, 1)$ such that
$\eta Q_{\max}^{\star} \leqslant \epsilon.$
If we ensure that $\big|\frac{Q_{iuv}(t)}{Q_{iuv}^{\star}} - 1\big| \leqslant \eta,$
for all $(i, u, v)$ and $t \geqslant 0$, then
\begin{equation*}
|Q_{iuv}(t) - Q_{iuv}^{\star}| = \big|\frac{Q_{iuv}(t)}{Q_{iuv}^{\star}} - 1\big| Q_{iuv}^{\star} \leqslant \eta Q_{\max}^{\star} \leqslant \epsilon,
\end{equation*}
and hence for all $t \geqslant 0$, $\|Q(t) - Q^{\star}\|_{\infty} \leqslant \epsilon.$
From the bounds above, we already know that
\begin{equation*}
\frac{Q_{iuv}(t)}{Q_{iuv}^{\star}} \in [\gamma(t) e^{-\sigma}, \gamma(t) e^{\sigma}],
\end{equation*}
for all $(i, u, v)$ and $t \geqslant 0$. Thus it suffices to make the interval $[\gamma(t) e^{-\sigma}, \gamma(t) e^{\sigma}]$ lie inside $[1 - \eta, 1 + \eta]$. 
To this end, choose small constants $\epsilon_s > 0$ and $\sigma_0 > 0$ such that $(1 - \epsilon_s) e^{-\sigma_0} \geqslant 1 - \eta, \quad (1 + \epsilon_s) e^{\sigma_0} \leqslant 1 + \eta.$
By Lemma~\ref{support_lemma4_secondary}, there exists $\delta_s > 0$ such that $\|Q(0) - Q^{\star}\|_{\infty} < \delta_s~\text{and~}|\gamma(t) - 1| < \epsilon_s$
for all $t \geqslant 0$. Moreover, since $\Phi$ is continuous at $Q^{\star}$ and $\Phi(Q^{\star}) = 0$, there exists $\delta_p > 0$ such that 
$\|Q(0) - Q^{\star}\|_{\infty} < \delta_p \text{~and~} \Phi(Q(0)) < \sigma_0.$
Set $\delta := \min\{\delta_s, \delta_p\}$. Then, if $\|Q(0) - Q^{\star}\| < \delta$, for all $t \geqslant 0$, $\gamma(t) \in (1 - \epsilon_s, 1 + \epsilon_s), \quad \sigma = \Phi(Q(0)) < \sigma_0,$
and hence $\frac{Q_{iuv}(t)}{Q_{iuv}^{\star}} \in [(1 - \epsilon_s) e^{-\sigma_0}, (1 + \epsilon_s) e^{\sigma_0}] \subset [1 - \eta, 1 + \eta]. $
As argued above, this implies $\|Q(t) - Q^{\star}\|_{\infty} \leqslant \epsilon$
for all $t \geqslant 0$. Therefore, $Q^{\star}$ is a Lyapunov stable equilibrium of $\dot{Q} = G(Q) - Q$. 
Finally, Lemma~\ref{support_lemma3_secondary} shows that $Q(t) \rightarrow Q^{\star}$ for every solution with $Q(0) > 0$. Thus, $Q^{\star}$ is both stable and globally attractive, i.e., it is a globally asymptotically stable equilibrium point of ODE $\dot{Q}(t) = G(Q(t)) - Q(t)$, and completes the proof.
\QED
\subsection{Proving $Q_n$ convergence to $Q^\star$}
\begin{lemma} \label{finite-positive}
Consider $(i, u, v) \in \X_0 \times \U \times \V$. Suppose $Q_0(i, u, v) > 0$ and $0 < a(\nu(i, u, v, n)) < 1$ for all $n$. Then emergency recursion in Eq.~\eqref{eq:Q} is well-defined for every finite $n$, and $Q_n(i ,u, v) > 0.$
Moreover, Eq.~\eqref{eq:Q} can be written as
\begin{eqnarray*}
&Q_{n+1}(i, u, v) = Q_n(i, u, v) + a(\nu(i, u, v, n))\nn\\
&\mathbb{1}_{\{X_n = i, U_n = u, V_n = v\}}\left(G_{iuv}(Q_n) - Q_n(i, u, v) + M_{n+1}(i, u, v)\right),\\
&\bE[M_{n+1}(i, u, v) | \pazocal{F}_n, X_n = i, U_n = u, V_n = v] = 0.
\end{eqnarray*}
where noise $M_{n+1}(i, u, v) := $
\begin{eqnarray*}
&\mathbb{1}_{\{X_{n+1} \in \X_0\}} \frac{\max_{v' \in \V} \min_{u' \in \U} Q_{n}(X_{n+1}, u', v')}{Q_n(i_0, u_0, v_0)} - G_{iuv}(Q_n). 
\end{eqnarray*}
\end{lemma}
\begin{proof}
We first prove positivity. Assume that $Q_n(i, u, v) > 0$ for every $(i, u, v) \in \X_0 \times \U \times \V$. If
$(X_n, U_n, V_n) \neq (i, u, v)$, then $Q_{n+1}(i, u, v) = Q_n(i, u, v) > 0$. If $(X_n, U_n, V_n) = (i, u, v)$, then Eq.~\eqref{eq:Q} gives
\begin{eqnarray*}
&Q_{n+1}(i, u, v) = (1 - a(\nu(i, u, v, n))) Q_n(i, u, v) \\
&+ a(\nu(i, u, v, n))\mathbb{1}_{\{X_{n+1} \in \X_0\}} \frac{\max_{v' \in \V} \min_{u' \in \U} Q_{n}(X_{n+1}, u', v')}{Q_n(i_0, u_0, v_0)}.
\end{eqnarray*}
The second term is nonnegative, and the first term is strictly positive, as $0 < a(\nu(i, u, v, n)) < 1$ and $Q_n(i, u, v) > 0$. Thus, $Q_{n+1}(i, u, v) > 0$. By induction, all components remain strictly positive at every finite time. Subsequently, conditional on $\pazocal{F}_n$ and $X_n = i$, $U_n = u$, $V_n = v$, we obtain
\begin{equation*}
G_{iuv}(Q_n) = \frac{e^{c_{uv}(i)}}{Q_n(i_0, u_0, v_0)} \sum_{j \in \X_0} P_{uv}^{\star}(j|i) \max_{v'}\min_{u'} Q_n(j, u', v').
\end{equation*}
Therefore $\mathbb{E}[M_{n+1}(i, u, v) | \pazocal{F}_n, X_n = i, U_n = u, V_n = v]$
\begin{align*}
&= \sum_{j \in \X_0} \bp(j | i, u, v) \frac{\max_{v' \in \V} \min_{u' \in \U} Q_{n}(j, u', v')}{Q_n(i_0, u_0, v_0)} - G_{iuv}(Q_n) \\
&=0.
\end{align*}
By substituting $\mathbb{1}_{\{X_{n+1} \in \X_0\}} \frac{\max_{v' \in \V} \min_{u' \in \U} Q_{n}(X_{n+1}, u', v')}{Q_n(i_0, u_0, v_0)} = G_{iuv}(Q_n) + M_{n+1}(i, u, v)$
into Eq.~\eqref{eq:Q} gives the stated stochastic-approximation form. This completes the proof.
\end{proof}
\begin{lemma}\label{local_Lipschitzness}
For $R > 0$ and $\delta > 0$, define
$$
D_{R, \delta} := \left\{Q > 0: \|Q\|_\infty \leqslant R, Q(i_0, u_0, v_0) \geqslant \delta \right\}.
$$
On $D_{R, \delta}$, $G$ is Lipschitz. Specifically, for all $Q, Q' \in D_{R, \delta}$,
$$
\|G(Q) - G(Q')\|_\infty \leqslant \left(\frac{1}{\delta} + \frac{R}{\delta^2}\right) \|Q - Q'\|_\infty.
$$
Consequently, $G(Q) - Q$ is Lipschitz on $D_{R, \delta}$. Furthermore, whenever $Q_n \in D_{R, \delta}$,
\begin{equation*}
\bE[|M_{n+1}(i, u, v)|^2 | \pazocal{F}_n,X_n = i, U_n = u, V_n = v] \leqslant \frac{4R^2}{\delta^2}.
\end{equation*}
\end{lemma}
\begin{proof}
For each $j \in \X_0$,
\begin{align*}
&|\max_{v'} \min_{u'} Q(j, u', v') - \max_{v'} \min_{u'} Q^{\prime}(j, u', v')|
\end{align*}
\begin{align*}
&\leqslant  \max_{v'} \max_{u'} |Q(j, u', v') - Q^{\prime}(j, u', v')| \leqslant \|Q - Q^{\prime}\|_{\infty}.
\end{align*}
For $(i, u, v) \in \X_0 \times \U \times \V$, we have $|G_{iuv}(Q) - G_{iuv}(Q')| $
\begin{align*}
&= \big|\frac{1}{Q(i_0, u_0, v_0)} \sum_{j \in \X_0} \bP(j|i, u, v) \max_{v'} \min_{u'} Q(j, u', v')\\
&- \frac{1}{Q'(i_0, u_0, v_0)} \sum_{j \in \X_0} \bP(j|i, u, v) \max_{v'} \min_{u'} Q'(j, u', v')\big| \\
&\leqslant \frac{1}{Q(i_0, u_0, v_0)} \sum_{j \in \X_0} \bP(j|i, u, v) |\max_{v'} \min_{u'} Q(j, u', v')\\
&- \max_{v'} \min_{u'} Q'(j, u', v')| + \big|\frac{1}{Q(i_0, u_0, v_0)} - \frac{1}{Q'(i_0, u_0, v_0)}\big|. \\
&\sum_{j \in \X_0} \bP(j|i, u, v) |\max_{v'} \min_{u'} Q'(j, u', v')| \\
&\leqslant \frac{1}{\delta} \|Q - Q'\|_{\infty} + \frac{R}{\delta^2} \|Q - Q'\|_{\infty} = \big(\frac{1}{\delta} + \frac{R}{\delta^2}\big) \|Q - Q'\|_{\infty}.
 \end{align*}
Taking the maximum over $(i, u, v)$ proves the Lipschitz bound. Since the drift is $G(q) - q$, it is also Lipschitz on $D_{R, \delta}$. For the noise bound, if $Q_n \in D_{R, \delta}$, then
\begin{eqnarray*}
&\big|\mathbb{1}_{\{X_{n+1} \in \X_0\}} \frac{\max_{v' \in \V} \min_{u' \in \U} Q_{n}(X_{n+1}, u', v')}{Q_n(i_0, u_0, v_0)}\big|\\
&\leqslant \frac{\|Q_n\|_{\infty}}{Q_n(i_0, u_0, v_0)} \leqslant \frac{R}{\delta}\text{~and}\\
&|G_{iuv}(Q_n)| = \big|\frac{1}{Q_n(i_0, u_0, v_0)} \sum_{j \in \X_0} \bP(j|i, u, v)\max_{v'}\min_{u'} \\
&Q_n(j, u', v')\big| \leqslant \frac{1}{\delta} \sum_{j \in \X_0} \bP(j|i, u, v) \|Q_n\|_{\infty} \leqslant \frac{R}{\delta}.
\end{eqnarray*}
Thus
\begin{eqnarray*}
&|M_{n+1}(i, u, v)| \leqslant \big|\mathbb{1}_{\{X_{n+1} \in \X_0\}} \frac{\max_{v' \in \V} \min_{u' \in \U} Q_{n}(X_{n+1}, u', v')}{Q_n(i_0, u_0, v_0)}\big|\\
&+ \big|G_{iuv}(Q_n)\big| \leqslant \frac{2R}{\delta},
\end{eqnarray*}
\begin{equation*}
\bE[|M_{n+1}(i, u, v)|^2 | \pazocal{F}_n, X_n = i, U_n = u, V_n = v] \leqslant \frac{4R^2}{\delta^2}.
\end{equation*}
This completes the proof.
\end{proof}

\begin{lemma}\label{Stability}
Suppose Assumption~\ref{Viable_Safe} holds. Under the positive initialization in Algorithm~\ref{algo:Q},
the Robbins-Monro stepsize conditions, and the sampling condition stated after Eq.~\eqref{eq:Q}, the emergency recursion satisfies
\begin{equation*}
\sup_{n \geqslant 0} \|Q_n\|_\infty < \infty, \quad \liminf_{n \rightarrow \infty} Q_n(i_0, u_0, v_0) > 0.
\end{equation*}
\end{lemma}
\begin{proof}
By Lemma~\ref{finite-positive}, $Q_n(i, u, v) > 0$ for every finite $n$ and every $(i, u, v) \in \X_0 \times \U \times \V$. Fix a sample path on which Assumption~\ref{Viable_Safe} holds. Thus, we have the conditional second moment bound.
We first prove the boundedness of an arbitrary component $(i, u, v) \in \X_0 \times \U \times \V$. Let $\{n_m^{iuv}\}_{m \geqslant 0}$ denote the sequential update times of component $(i, u, v)$ after $N_H$, and define $\bar{Q}_m^{iuv} := Q_{n_m^{iuv}}(i, u, v)$ and $\alpha_m^{iuv} := a(\nu(i, u, v, n_m^{iuv})).$
Along this local clock, Eq.~\eqref{eq:Q} yields
\begin{eqnarray*}
&\bar{Q}_{m+1}^{iuv} = (1 - \alpha_m^{iuv}) \bar{Q}_m^{iuv} + \alpha_m^{iuv} \mathbb{1}_{\{X_{n_m^{iuv}+1} \in \X_0\}} \\
&\cdot\frac{\max\limits_{v' \in \V} \min\limits_{u' \in \U} Q_{n_m^{iuv}}(X_{n_m^{iuv}+1}, u', v')}{Q_{n_m^{iuv}}(i_0, u_0, v_0)}.
\end{eqnarray*}
Set 
$\bar{H}_m^{iuv} := \bE\big[\mathbb{1}_{\{X_{n_m^{iuv}+1} \in \X_0\}} \frac{\max\limits_{v' \in \V} \min\limits_{u' \in \U} Q_{n_m^{iuv}}(X_{n_m^{iuv}+1}, u', v')}{Q_{n_m^{iuv}}(i_0, u_0, v_0)} |$ $\pazocal{F}_{n_m^{iuv}}\big],$ and $\xi_{m+1}^{iuv} := \mathbb{1}_{\{X_{n_m^{iuv}+1} \in \X_0\}} \frac{\max\limits_{v' \in \V} \min\limits_{u' \in \U}Q_{n_m^{iuv}}(X_{n_m^{iuv}+1}, u', v')}{Q_{n_m^{iuv}}(i_0, u_0, v_0)}$ $- \bar{H}_m^{iuv}.$
Then $\bE[\xi_{m+1}^{iuv} | \pazocal{F}_{n_m^{iuv}}] = 0$. Given that the sample target is nonnegative, Jensen's inequality and Assumption~\ref{Viable_Safe} gives $0 \leqslant \bar{H}_m^{iuv} \leqslant \big(\bE\big[\big(\mathbb{1}_{\{X_{n_m^{iuv}+1} \in \X_0\}} $
\begin{equation*}
\frac{\max_{v' \in \V} \min_{u' \in \U} Q_{n_m^{iuv}}(X_{n_m^{iuv}+1}, u', v')}{Q_{n_m^{iuv}}(i_0, u_0, v_0)}\big)^2 | \pazocal{F}_{n_m^{iuv}}\big]\big)^{1/2} \leqslant C_H.
\end{equation*}
Furthermore, $\bE[(\xi_{m+1}^{iuv})^2 | \pazocal{F}_{n_m^{iuv}}]\leqslant C_H^2.$
Define
\begin{align*}
y_{m+1}^{iuv} &= (1 - \alpha_m^{iuv}) y_m^{iuv} + \alpha_m^{iuv} C_H, \quad y_0^{iuv} = \bar{Q}_0^{iuv}, \\
w_{m+1}^{iuv} &= (1 - \alpha_m^{iuv}) w_m^{iuv} + \alpha_m^{iuv} \xi_{m+1}^{iuv}, \quad w_0^{iuv} = 0.
\end{align*}
Given that $C_H$ is the fixed point of the recursion for $y_m^{iuv}$, let $e_m^{iuv} := y_m^{iuv} - C_H$. Then
\begin{eqnarray*}
e_{m+1}^{iuv} = (1 - \alpha_m^{iuv}) e_m^{iuv} \text{~and~} e_m^{iuv} = e_0^{iuv} \prod_{\ell = 0}^{m-1} (1 - \alpha_\ell^{iuv}).
\end{eqnarray*}
Utilizing the inequality $\log (1 - x) \leqslant - x$ on the interval $[0, 1)$, 
\begin{equation*}
\prod_{\ell = M}^{m - 1} (1 - \alpha_\ell^{iuv}) = \exp \big(\sum_{\ell=M}^{m-1} \log (1 - \alpha_\ell^{iuv})\big) \leqslant \exp \big(- \sum_{\ell = M}^{m-1} \alpha_\ell^{iuv}\big)
\end{equation*}
Given that $\sum_{m} \alpha_m^{iuv} = \infty$, it follows that $\exp \big(- \sum_{\ell = M}^{m-1} \alpha_\ell^{iuv}\big) \rightarrow 0$. Thus, the entire product converges to zero. As a result, $y_m^{iuv} - C_H = e_m^{iuv} \rightarrow 0$, and therefore $y_m^{iuv} \rightarrow C_H.$
For $w_m^{iuv}$, using the martingale-difference property and the conditional second-moment bound, $\bE[(w_{m+1}^{iuv})^2 | \pazocal{F}_{n_m^{iuv}}]$
\begin{align*}
 &= (1 - \alpha_m^{iuv})^2 (w_m^{iuv})^2 + (\alpha_m^{iuv})^2
\bE[(\xi_{m+1}^{iuv})^2 | \pazocal{F}_{n_m^{iuv}}] \\
&\leqslant (1 - \alpha_m^{iuv}) (w_m^{iuv})^2 + (\alpha_m^{iuv})^2 C_H^2 \\
&= (w_m^{iuv})^2 + (\alpha_m^{iuv})^2 C_H^2 - \alpha_m^{iuv} (w_m^{iuv})^2.
\end{align*}
Given that $\sum_m (\alpha_m^{iuv})^2 C_H^2 = C_H^2 \sum_m (\alpha_m^{iuv})^2 < \infty$, the Robbins-Siegmund almost-supermartingale theorem implies that $(w_m^{iuv})^2$ converges almost surely and
\begin{equation*}
\sum_m \alpha_m^{iuv} (w_m^{iuv})^2 < \infty \quad \text{almost surely}.
\end{equation*}
Since $\sum_m \alpha_m^{iuv} = \infty$, the only possible limit is zero. Therefore
$w_m^{iuv} \rightarrow 0 \quad\text{almost surely}.$
Set $\Delta_m^{iuv} := \bar{Q}_m^{iuv} - y_m^{iuv} - w_m^{iuv}$. By subtracting the recursions, we obtain
\begin{equation*}
\Delta_{m+1}^{iuv} = (1 - \alpha_m^{iuv}) \Delta_m^{iuv} + \alpha_m^{iuv} (\bar{H}_m^{iuv} - C_H).
\end{equation*}
Given that $\bar{H}_m^{iuv} \leqslant C_H$ and $\Delta_0^{iuv} = 0$, induction proves that $\Delta_m^{iuv} \leqslant 0$ for all $m$. Thus, $\bar{Q}_m^{iuv} \leqslant y_m^{iuv} + w_m^{iuv}.$
The right-hand side is bounded as $y_m^{iuv} \rightarrow C_H$ and $w_m^{iuv} \rightarrow 0$. Therefore, the local sequence $\{\bar{Q}_m^{iuv}\}$ is bounded above. Between two update times of component $(i, u, v)$, the component $Q_n(i, u, v)$ does not change. Consequently, $Q_n(i, u, v)$ is bounded above for every $(i, u, v)$. Given that $\X_0 \times \U \times \V$ is finite, it follows that
$\sup_{n \geqslant 0} \|Q_n\|_\infty < \infty.$
It remains to prove separation from the normalization boundary. Let $\{n_m^{0}\}_{m \geqslant 0}$ denote the sequential update times of the component $(i_0, u_0, v_0)$ after $N_H$, and define $\bar{r}_m := Q_{n_m^{0}}(i_0, u_0, v_0), \quad \alpha_m^0 := a(\nu(i_0, u_0, v_0, n_m^{0})).$
Along the anchor local clock,
\begin{eqnarray*}
&\bar{r}_{m+1} = (1 - \alpha_m^0) \bar{r}_m + \alpha_m^0 \mathbb{1}_{\{X_{n_m^0 + 1} \in \X_0\}}\\
&\frac{\max_{v' \in \V} \min_{u' \in \U} Q_{n_m^0}(X_{n_m^0 + 1}, u', v')}{Q_{n_m^0}(i_0, u_0, v_0)}.
\end{eqnarray*}
Define
\begin{eqnarray*}
\bar{H}_m^0 &:= \bE\big[\mathbb{1}_{\{X_{n_m^0 + 1} \in \X_0\}} \frac{\max\limits_{v' \in \V} \min\limits_{u' \in \U} Q_{n_m^0}(X_{n_m^0 + 1}, u', v')}{Q_{n_m^0}(i_0, u_0, v_0)} | \pazocal{F}_{n_m^0}\big], \\
\xi_{m+1}^0 &:= \mathbb{1}_{\{X_{n_m^0 + 1} \in \X_0\}} \frac{\max\limits_{v' \in \V} \min\limits_{u' \in \U} Q_{n_m^0}(X_{n_m^0 + 1}, u', v')}{Q_{n_m^0}(i_0, u_0, v_0)} - \bar{H}_m^0.
\end{eqnarray*}
By Assumption~\ref{Viable_Safe}, $\bar{H}_m^0 \geqslant c_H$. The same conditional second-moment argument as above yields $w_m^0 \rightarrow 0 \quad \text{almost surely},$
where $w_{m+1}^0 = (1 - \alpha_m^0) w_m^0 + \alpha_m^0 \xi_{m+1}^0, \quad w_0^0 = 0.$
Define
\begin{equation*}
y_{m+1}^0 = (1 - \alpha_m^0) y_m^0 + \alpha_m^0 c_H, \quad y_0^0 = \bar{r}_0.
\end{equation*}
Given that $\sum_m \alpha_m^0 = \infty$, it follows that $y_m^0 \rightarrow c_H$. Now set $\Delta_m^0 := \bar{r}_m - y_m^0 - w_m^0$. By subtracting the recursions, 
\begin{equation*}
\Delta_{m+1}^0 = (1 - \alpha_m^0) \Delta_m^0 + \alpha_m^0 (\bar{H}_m^0 - c_H).
\end{equation*}
Given that $\bar{H}_m^0 \geqslant c_H$ and $\Delta_0^0 = 0$, induction proves that $\Delta_m^0 \geqslant 0$ for all $m$. Consequently, $\bar{r}_m \geqslant y_m^0 + w_m^0.$
By taking lower limits and using $y_m^0 \rightarrow c_H$ and $w_m^0 \rightarrow 0$, $\liminf_{m \rightarrow \infty} \bar{r}_m \geqslant c_H > 0.$
Between two anchor update times, $Q_n(i_0, u_0, v_0)$ is unchanged. Therefore, $\liminf_{n \rightarrow \infty} Q_n(i_0, u_0, v_0) > 0.$ 
Combining this boundary separation with the proven boundedness yields
\begin{equation*}
\sup_{n \geqslant 0} \|Q_n\|_\infty < \infty, \quad \liminf_{n \rightarrow \infty} Q_n(i_0, u_0, v_0) > 0 \quad \text{almost surely}.
\end{equation*}
This completes the proof.
\end{proof}

\noindent{\em Proof of Theorem~\ref{emergency_convergence}:}
By Lemma~\ref{finite-positive}, Eq.~\eqref{eq:Q} can be written as the asynchronous stochastic-approximation recursion
\begin{eqnarray*}
&Q_{n+1}(i, u, v) = Q_n(i, u, v) + a(\nu(i, u, v, n)) \\
&\mathbb{1}_{\{X_n = i, U_n = u, V_n = v\}}\left(G_{iuv}(Q_n) - Q_n(i, u, v) + M_{n+1}(i, u, v)\right),\\
&\bE[M_{n+1}(i, u, v) | \pazocal{F}_n, X_n = i, U_n = u, V_n = v] = 0. 
\end{eqnarray*}
By Lemma~\ref{Stability}, the emergency iterates are almost surely bounded and asymptotically far from the normalization boundary. Hence, on almost every sample path, there exist finite random constants $R < \infty$, $\delta > 0$, and $N < \infty$ such that
\begin{equation*}
\|Q_n\|_\infty \leqslant R, \quad Q_n(i_0, u_0, v_0) \geqslant \delta, \text{~for~} n \geqslant N.
\end{equation*}
 On the set
$D_{R, \delta} := \left\{Q > 0: \|Q\|_\infty \leqslant R, Q(i_0, u_0, v_0) \geqslant \delta\right\},$
Lemma~\ref{local_Lipschitzness} shows that the drift $G(Q) - Q$ is Lipschitz and the martingale-difference noise has uniformly bounded conditional second moment. 
Utilizing the standard localization argument, we may modify the drift and noise outside $D_{R, \delta}$, ensuring that the localized recursion has globally Lipschitz drift while remaining consistent with Eq.~\eqref{eq:Q} for all $n \geqslant N$ on the fixed sample path. By applying Theorem~2.5 from \cite{borkar2000ode} to this localized asynchronous recursion and utilizing Theorem~\ref{Risk:Thm3} for the limiting ODE $\dot Q(t) = G(Q(t)) - Q(t)$, we conclude that $Q_n \rightarrow Q^\star \quad\text{almost surely}.$
Taking the reference coordinate and using the normalization $Q^\star(i_0, u_0, v_0) = \Lambda^\star$ results in
\begin{equation*}
Q_n(i_0, u_0, v_0) \rightarrow \Lambda^\star \quad\text{almost surely}.
\end{equation*}
This completes the proof.
\QED
\section*{References}
\vspace{-7 mm}

\end{document}